\documentclass{article}
\usepackage{graphicx} 

\usepackage{microtype}
\usepackage{pgfplots}
\usepackage{lipsum}		
\usepackage{booktabs} 
\usepackage{makecell} 
\usepackage[
    top=3cm,
    bottom=3cm,
    left=2cm,
    right=2cm,
]{geometry}
\usepackage{setspace}
\usepackage{comment}
\usepackage{doi}
\usepackage{multirow}
\usepackage{multicol}
\usepackage{xcolor}
\usepackage{subcaption}
\usepackage[normalem]{ulem}
\usepackage{amsmath,amsfonts,bm}
\usepackage{amssymb,amsthm}
\usepackage{enumitem}
\usepackage{subcaption}
\usepackage{adjustbox}
\usepackage{afterpage}
\usepackage{threeparttable}
\usepackage[skip=10pt]{parskip}  
\usepackage{fancyhdr}
\usepackage{CJKutf8}  
\usepackage{CJK}
\usepackage{placeins} 

\usepackage{float} 

\usepackage[
    backend=biber,
    mincitenames=1,
    maxcitenames=1,
    maxbibnames=3,
    sortcites=true
]{biblatex}
\newcolumntype{C}[1]{>{\centering\let\newline\\\arraybackslash\hspace{0pt}}m{#1}}

\title{BlueLM-2.5-3B Technical Report}
\author{vivo AI Lab}
\date{July 2025}

\begin{document}

\maketitle

\renewenvironment{abstract}{%
  \small
  \begin{center}%
    \bfseries \abstractname 
  \end{center}%
  \list{}{%
    \setlength{\leftmargin}{20pt}
    \setlength{\rightmargin}{20pt}
    \item[]%
  }\item\relax
}{%
  \endlist
}

\begin{abstract}
We present \textbf{BlueLM-2.5-3B}, a compact and unified dense Multimodal Large Language Model (MLLM) designed for efficient edge-device deployment, offering strong general-purpose and reasoning capabilities. To the best of our knowledge, this is the first 3B-scale MLLM to support both \textit{thinking} and \textit{non-thinking} modes, while also enabling explicit control over thinking token budget. BlueLM-2.5-3B is developed through diversified data curation, key data resampling, hybrid heterogeneous reinforcement learning, and a high-performance training infrastructure. Our model achieves superior multimodal capacity while preserving competitive pure-text performance with only \textbf{2.9 billion parameters}. We conduct comprehensive evaluations  across a broad range of multimodal and text-only benchmarks. In thinking mode, BlueLM-2.5-3B achieves comparable performance to Qwen3-4B on text-only benchmarks, and trails the larger Kimi-VL-A3B-16B by only about 5\% on average across multimodal evaluations. In non-thinking mode, it outperforms Qwen2.5-VL-3B on the majority of multimodal benchmarks. Additionally, BlueLM-2.5-3B exhibits exceptional data efficiency. All of the aforementioned performance is achieved with substantially less total training data than Qwen2.5-VL-3B and Qwen3-4B. 
We hope our work contributes to the advancement of high-performance, on-device MLLMs and provides meaningful insights to the research community.
\end{abstract}

\begin{figure}[h]
    \centering
    \label{fig0}
    \includegraphics[width=0.9\linewidth]{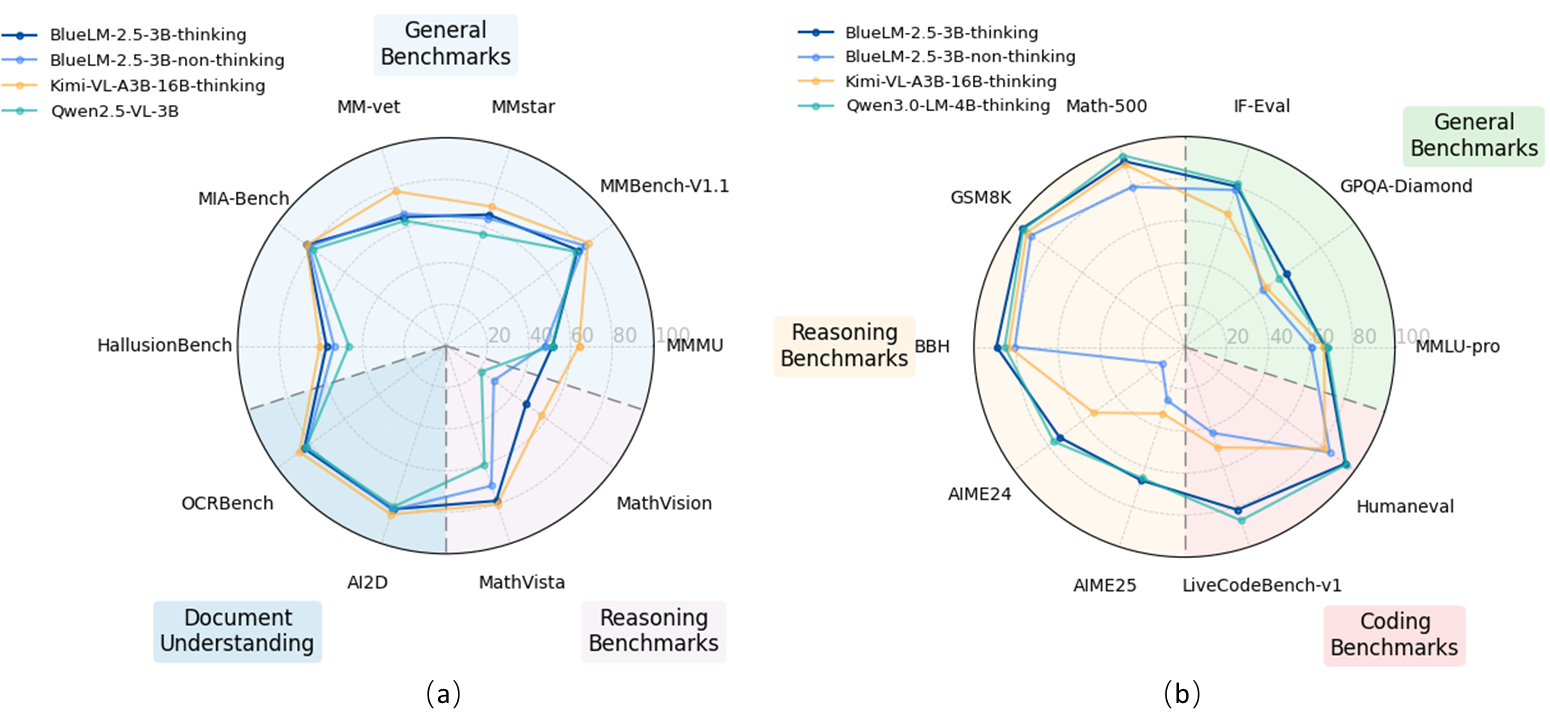}
    \caption{Multimodal(a) and Pure-text(b) Benchmark Performance of BlueLM-2.5-3B.}
\end{figure}

\thispagestyle{fancy}

\newpage
\begin{spacing}{1.2}
    \tableofcontents
\end{spacing}
\newpage

\section{Introduction}

Progress in artificial intelligence has been profoundly shaped by the rise of Multimodal Large Language Models(MLLMs), which serve as the core infrastructure for multimodal systems. These models facilitate sophisticated interactions between artificial agents and their surroundings by processing visual data, interpreting composite information streams~\cite{yue2024mmmu}, and executing tasks across digital~\cite{xie2024osworld, openai2025computeruse} and physical domains ~\cite{brohan2023rt, black2410pi0}. The integration of visual and linguistic representations within a single architecture has spurred transformative applications, ranging from complex inference tasks~\cite{bai2025qwen25vltechnicalreport, team2025gemini, openai2024gpt4ocard} and visual content manipulation~\cite{kampf2025experiment} to autonomous navigation systems~\cite{xu2024drivegpt4, tian2024drivevlm, pan2024vlp} and embodied intelligence~\cite{figure2024helix, kim2024openvla}. 
However, the high computational cost of large-scale LLMs and MLLMs limits their current deployment to cloud servers only, excluding edge-side applications like mobile devices, vehicles, and robotics. This reliance on cloud infrastructure raises concerns about energy consumption, latency, and privacy, particularly in offline or security-sensitive scenarios. 

Recent advances in lightweight model design, supported by improved training data and knowledge distillation, enable efficient edge-side deployment~\cite{abdin2024phi, bai2025qwen25vltechnicalreport, yang2025qwen3}. Compact MLLMs~\cite{team2025gemini,abdin2024phi, beyer2024paligemma, wu2024deepseekvl2mixtureofexpertsvisionlanguagemodels}offer practical advantages, including lower latency, offline usability, and enhanced privacy, making them viable for ubiquitous edge devices with growing computational capabilities. However, building powerful end-side MLLMs poses challenges due to strict constraints on model size and inference computing resources. This demands more meticulous architecture choices and training methods to maximize the performance of end-side MLLMs.

Currently, the development of reasoning MLLMs capable of incorporating deeper and longer reasoning on multimodal inputs and thereby tackling more complex problems in the multimodal domain has brought in a new era. However, small multimodal models with strong capabilities of simultaneous multimodal reasoning and general multimodal understanding are still significantly lacking. 

In this work, we present BlueLM-2.5-3B, the first edge-side multimodal model that combines thinking and non-thinking capabilities in a single model, which is capable of adaptively switching between the two thinking modes based on user query types or chat templates. We achieve this flexibility through systematic innovation in three key aspects: \textbf{training data}, \textbf{training algorithms}, and \textbf{training infrastructure}. Regarding training data curation, we increase the proportion of pure text samples in the pure-text and multimodal data joint pre-training to 40\%, thereby effectively mitigating the text capability degradation issue. Also, a substantial amount of 3.3T data was synthesized for integrated image-text reasoning tasks, leading comparable models in terms of data scale and quality. In terms of training algorithms, we propose an innovative multi-phase training pipeline to effectively optimize model performance across both general and reasoning tasks. Initially, text pre-training utilizes large-model distillation to enhance small-model accuracy (+4\%) while reducing costs. In the MLLM pre-training phase, increasing text data proportions streamlines training by eliminating separate MLP + ViT stages. Then, reasoning-enhanced training adopts curriculum-inspired multi-stage distillation, blending pre-training data with synthesized reasoning data to strengthen complex reasoning and contextual understanding. After that, the fast decay phase progressively reduces batch size while prioritizing high-quality long-text data, boosting long-context reasoning. During supervised fine-tuning (SFT), key data resampling strategy further enhances complex reasoning performance, with a flexible "think" switch in the instruction data to balance exploration and structured responses. Finally, we design a hybrid reinforcement learning approach to improve reward generalization, utilizing different reward forms according to the specific task types. We also introduces a meticulously designed length penalty strategy that allows the model to perform robustly under tight token budgets, thus better accommodating the latency demands of edge-side deployment. 
Regarding training infrastructure, context parallelism for long-sequence pre-training improved training efficiency by 1.66×. Load balancing and multi-sample batching raised model utilization to 45\%, while asynchronous RL training improved efficiency by 1.32×. Additionally, inference scheduling optimizations reduced idle time, boosting RL performance by another 1.3×.
 
With a model size of only 2.9B, sufficient evaluation results show that our BlueLM-2.5-3B is comparable to several cutting-edge open-source models of larger sizes across multiple benchmarks, highlighting both its efficiency and effectiveness. Notably, in the thinking mode, compared to the multimodal SOTA kimi-VL-A3B-16B, our model's performance on all of the 10 multimodal benchmarks are only about 5\% behind in average. Moreover, compared to the latest small-scale pure-text SOTA Qwen3-4B, our model's scores on the pure-text benchmarks are comparable with only 60\% of the language model's parameters. Furthermore, in both of the multimodal and text reasoning benchmarks, our model significantly outperforms Qwen2.5-VL-72B, a much larger cutting-edge multimodal large language model. In the non-thinking mode, BlueLM-2.5-3B outperforms Qwen2.5-VL-3B, another cutting-edge VLM with 30\% more parameters, in majority of the multimodal benchmarks.

By refining the data processing pipeline, meticulously designing training strategies, and enhancing training efficiency, we show that integrating adaptive thinking ability into small multimodal models on edge devices is feasible. We hope our work can inspire further research advancements in the field of edge-based multimodal models.

\FloatBarrier

\section{Approach}

\subsection{Model Architecture}

\begin{figure}[H]
    \centering
    \includegraphics[width=0.7\linewidth]{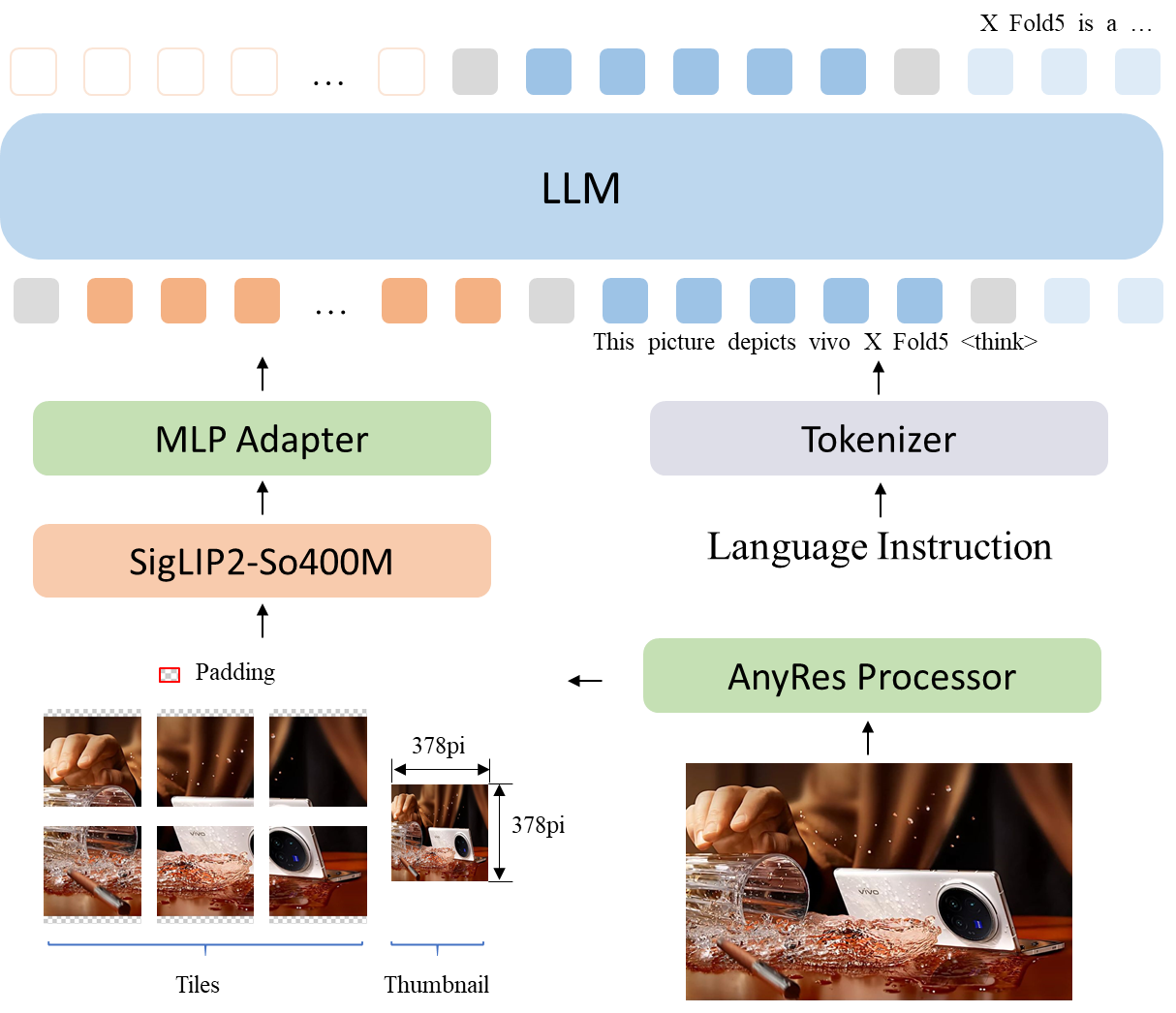}
    \caption{Model architecture of BlueLM-2.5-3B.}
\label{fig:blue_vit}
\end{figure}

As illustrated in Figure~\ref{fig:blue_vit}, the architecture of BlueLM-2.5-3B comprises three main components: a Vision Transformer (ViT), an adapter module, and a dense Large Language Model (LLM). Each component is described in detail below.

\paragraph{Vision Transformer (ViT)}
We adopt SigLIP2 \cite{tschannen2025siglip}  (so400m-patch14-384) to initialize the ViT(also referred to as vision encoder), which consists of 27 transformer layers and contains approximately 400 million parameters. While the nominal input resolution of SigLIP2 is 384×384, the patch size of 14 does not evenly divide the input dimensions, resulting in an effective receptive field of 378×378. Utilizing the nominal resolution would result in information loss at the image boundaries. To preserve the complete visual context, we adjust the input resolution to 378×378. 

To enable dynamic input resolutions, we integrate the AnyRes~\cite{DBLP:conf/cvpr/LiuLLL24} Processor, which maintains the original aspect ratio of the input image via minimal scaling and applies additional padding if necessary. The resulting image is then partitioned into non-overlapping 378×378 tiles and processed using a simple 4×4 grid layout. This design allows the model to support input images with resolutions up to 1512×1512 pixels.

A key advantage of the AnyRes strategy lies in its computational efficiency. As illustrated in Figure~\ref{fig:vit_latency}, the inference time of the ViT in Qwen2.5-VL increases quadratically with the input resolution. In contrast, our tile-based approach enables parallel processing across tiles, substantially reducing inference latency. Moreover, the fixed-length token representation simplifies deployment on resource-constrained devices, such as smartphone chips, by facilitating consistent memory allocation and efficient execution.

\pgfplotsset{compat=newest}

\begin{figure}[ht]
\centering
\resizebox{0.6\textwidth}{!}{
\begin{tikzpicture}
\centering
\begin{axis}[
    width=12cm,
    height=7.5cm,
    xlabel={Token Number (k)},
    ylabel={Inference Latency (ms)},
    title={\textbf{ViT Inference Latency vs Token Number}},
    legend pos=north west,
    grid=both,
    xtick={1,2,3,4,5},
    ymin=0,
    ymajorgrids=true,
    tick label style={font=\small},
    label style={font=\small},
    title style={font=\small},
    legend style={font=\small},
]

\addplot[
    color=blue,
    mark=*,
    thick,
] coordinates {
    (1,256)
    (2,354)
    (3,482)
    (4,610)
    (5,708)
};
\addlegendentry{BlueLM-2.5-3B}

\addplot[
    color=orange!90!red,
    mark=square*,
    thick,
] coordinates {
    (1,200)
    (2,542)
    (3,888)
    (4,1548)
    (5,2395)
};
\addlegendentry{Qwen2.5-VL}

\end{axis}
\end{tikzpicture}
}
\caption{ViT Inference Latency vs Token Number.}
\label{fig:vit_latency}
\end{figure}
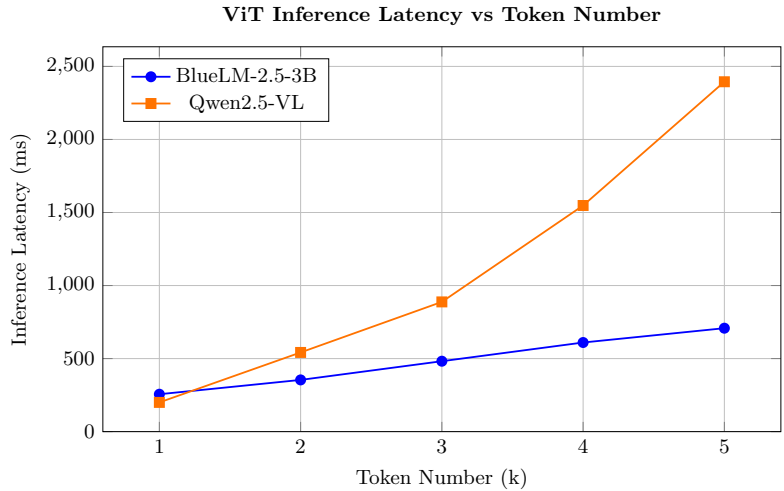

\paragraph{Adapter Module}
To align visual representations with the language model input space, we utilize a two-layer multilayer perceptron (MLP) as the adapter module(also referred to as projection module). The MLP maps the image token space to the embedding space of the LLM. Additionally, a 2×2 spatial downsampling is applied to reduce the token length and enhance computational efficiency.

\paragraph{LLM Module}
For the language modeling backbone, we trained an in-house developed LLM, a dense transformer-based model with 2.5 billion parameters. The in-house LLM is optimized for multimodal alignment, providing strong semantic understanding and reasoning capabilities. 

Overall, our model prioritizes compactness and efficiency by design. As shown in Table~\ref{tab:model_parameter_compare}, BlueLM-2.5-3B contains 22\% fewer parameters than comparable models like Qwen2.5-VL-3B. 

\begin{table}[ht]
    \centering
    \caption{Model Parameter Comparison}
    \renewcommand{\arraystretch}{1.2} 
    \resizebox{0.9\textwidth}{!}{
    \begin{tabular}{l C{5cm} C{5cm} C{5cm}}
        \toprule
        \textbf{Model} & \textbf{ViT Parameters} & \textbf{LLM Parameters} & \textbf{Total Parameters} \\
        \midrule
        Qwen2.5-VL-3B & 0.67B & 3.09B & 3.76B \\
        Kimi-VL-A3B-16B & 0.4B & 16B / A2.8B & 16.4B / A3.2B \\
        \midrule
        \textbf{BlueLM-2.5-3B} & 0.4B & 2.5B & 2.9B \\
        \bottomrule
    \end{tabular}
    }
    \label{tab:model_parameter_compare}
\end{table}

\FloatBarrier

\subsection{Pre-training Stages}


The full pre-training process of our multimodal model consists of a pure-text pre-training phase followed by a multimodal pre-training phase. The training of multimodal models is carried out in three stages: joint general pre-training stage, reasoning-enhanced stage, and joint fast-decay and long-context activation stage. The detailed configurations for each stage are presented in Table~\ref{tab:joint_pretraining_stages}.

\begin{table}[ht]
    \centering
    \caption{Joint Pre-training Stage Parameters}
    \renewcommand{\arraystretch}{1.2} 
    \resizebox{0.98\textwidth}{!}{
    \begin{tabular}{l C{5cm} C{5cm} C{6cm}}
    \toprule
         & \textbf{Joint}\newline \textbf{Pre-training Stage} & \textbf{Reasoning}\newline \textbf{Enhanced Stage} & \textbf{Joint Fast-decay \& Long-}\newline \textbf{context Activation Stage} \\
        \midrule \midrule
        \textbf{Tokens} & 4T & 2.5T & 1.3T \\
        \midrule
        \textbf{Sequence Length} & 4096 & 4096 & 32768 \\
        \midrule
        \textbf{Peak LR} & 2e-4 & - & - \\
        \midrule
        \textbf{Batch Size} & 64M & 64M & 64M / 16M \\
        \midrule
        \textbf{Training} & ViT \& LLM & ViT \& LLM & ViT \& LLM \\
        \midrule
        \textbf{Data Type} & Text \& Captioning \newline OCR \& VQA \newline GUI Agent & High-quality Text\newline High-quality Multimodal\newline Synthesis QA & Long Text\newline Long Multimodal \\
        \bottomrule
    \end{tabular}
    }
    \label{tab:joint_pretraining_stages}
\end{table}

\subsubsection{Pure-Text Pre-training Stage}
To initialize the LLM parameters of BlueLM-2.5-3B, we first train a text-only model LLM-base-3B. LLM-base-3B is a chat model that demonstrates comparable capabilities to Qwen2.5-3B-Instruct~\cite{bai2025qwen25vltechnicalreport} across language understanding, knowledge, reasoning, mathematics, coding, and instruction following.
The training of LLL-base-3B comprises three stages: general pre-training, fast decay (FD), and fine-tuning. Both the pre-training and FD stages utilize model distillation based on logits alignment. The teacher models are a 7B pre-trained model and a 7B FD model, respectively.
The 7B pre-trained model was pre-trained from scratch on 6T tokens of pure text data with a maximum sequence length of 4096. The LLM-base-3B was derived via pruning~\cite{muralidharan2024compact} from the 7B pre-trained model, followed by further training on an additional 3T tokens of distinct pure text data, also with a maximum sequence length of 4096. Both the 7B FD model and the 3B FD model were trained using the same 300B tokens of data.
Compared to training a 3B pure-text model from scratch, the approach based on pruning and distillation from 7B teacher models demonstrates consistent improvements across various benchmark tasks.

\begin{figure}
    \centering
    \includegraphics[width=0.9\linewidth]{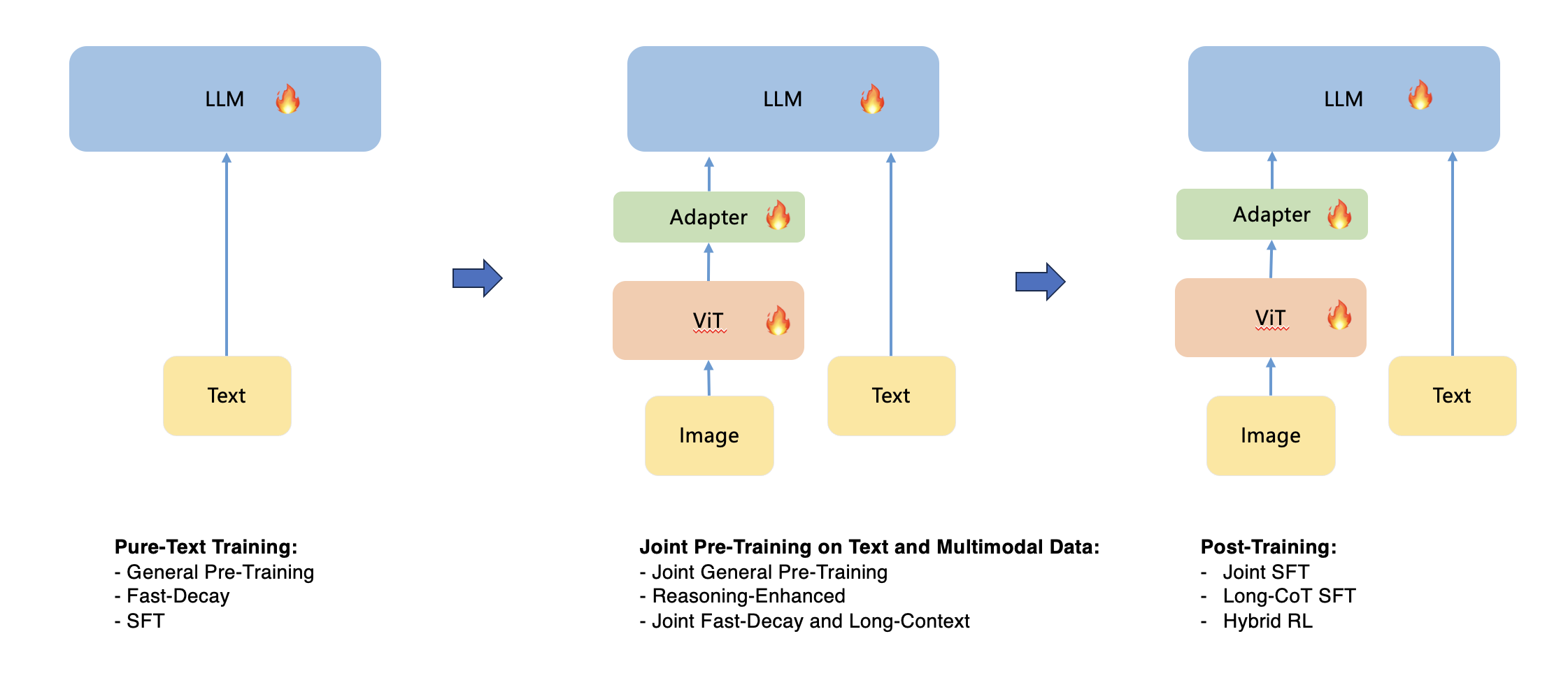}
    \caption{Training stages of BlueLM-2.5-3B.}
    \label{fig:training_strategy}
\end{figure}

\subsubsection{Joint General Pre-training Stage}
Joint general training comprises two sub-stages. In sub-stage 1, the ViT and LLM are frozen. Only the adapter layer is trained using 2.5M high-quality image-text pairs. In sub-stage 2, all parameters are unfrozen (Figure~\ref{fig:training_strategy}) and jointly trained on a combined dataset of 4T tokens, comprising multi-type, multi-task image-text data and pure text data in a 6:4 ratio. Due to the substantial inclusion of text data, a separate pre-training phase for the visual encoder and adapter layer was omitted. To maximize GPU utilization and achieve a near-linear speedup, we employed a large batch size of 64M tokens. To maintain performance at this scale, the learning rate was scaled up~\cite{mccandlish2018empirical}, while ensuring the maximum learning rate remained below the typical maximum fine-tuning learning rate for the language model to mitigate significant catastrophic forgetting. All MLLM pre-training stages share a single large learning rate decay schedule (12T tokens) to avoid performance degradation caused by repeated learning rate warmup and decay. Layer-wise learning rate decay is applied to the visual encoder to preserve features in its shallower layers, enhancing training stability and performance.
Both stages utilize a maximum sequence length of 4096.

\FloatBarrier

\subsubsection{Reasoning-Enhanced Stage}
To further enhance the model’s reasoning capabilities, we implemented reasoning-focused continued pre-training, targeting both textual and multimodal reasoning tasks. For the textual modality, we enriched the training corpus by increasing the proportion of STEM-related content, programming code, general reasoning examples, and synthetic data. The incorporation of high-quality reasoning-oriented pretraining datasets led to moderate improvements in general reasoning performance~\cite{shao2024deepseekmath}, whereas synthetic data demonstrated significant gains across a variety of downstream tasks~\cite{yang2025qwen3, abdin2024phi4}. On the multimodal side, we introduced a substantial amount of high-quality vision-language question answering (VQA) data that involves reasoning tasks, aiming to strengthen the model’s cross-modal reasoning capability.~\cite{team2025kimivl}
In this phase, the model was trained on approximately 2.5T high-quality tokens, including 1.3T  textual tokens and 1.2T multimodal tokens, using a maximum sequence length of 4096. Additionally, the synthetic data includes not only traditional short chain-of-thought (CoT) examples but also long CoT sequences—up to 4096 tokens in length—spanning both textual and multimodal reasoning tasks. This design aims to improve the model’s ability to handle complex, multi-step reasoning processes.~\cite{yue2025does}



\subsubsection{Joint Fast-decay and Long-context Activation Stage}
To further enhance the model’s long-context capabilities and reasoning performance, we conducted an incremental training strategy. Immediately following the initial reasoning-enhanced stage, to further improve the model’s long-context extrapolation capability, we adjusted the positional encoding from RoPE~\cite{su2024roformer} to YaRN~\cite{peng2023yarn} and expanded the model’s context length from 4k to 32k through fast-decay training. This phase utilized approximately 0.7T training tokens, comprising data with a distribution similar to that of the pretraining stage, as well as high-quality native long-text data and long-reasoning data. Subsequently, we carried out an additional 0.6T-token continued pretraining with a 32k context length. During this process, we reduced the global batch size and increased the proportion of native long-text and high-quality long-reasoning data to over 80\%. Our experiments demonstrated that replaying and increasing the share of high-quality long-reasoning data effectively improves the model’s long-form reasoning capabilities.

\subsection{Post-training Stages}

\subsubsection{Joint Supervised Fine-Tuning with Thinking Mode Fusion}
In this phase, we fine-tune the model obtained during the pretraining stage. We conduct joint fine-tuning on text and multimodal instructions while integrating control over reasoning modes. For various tasks, we construct datasets in both long CoT and standard CoT formats, performing quality filtering based on instruction requirements. Additionally, we introduce a special token,  [\textbar BlueThink\textbar], as a control switch, enabling the model to adopt different reasoning modes depending on the task scenario. During fine-tuning, both the special token and format query labels are masked to exclude them from loss computation. In multi-turn dialogues, the reasoning mode can be deactivated simply by omitting the reasoning token in the current dialogue.

Since our base model inherently supports a 32k context length, we employ full-length 32k fine-tuning. For standard CoT data, we train for 3 epochs, whereas for long CoT data, we extend training to 9 epochs. The learning rate follows a warmup schedule from 0 to 1e-5, followed by a cosine decay.

\subsubsection{Long-CoT Supervised Fine-Tuning}
In this work, we construct a high-quality dataset targeted at long-form reasoning tasks. The dataset primarily covers complex queries from contests related to STEM. Each query is accompanied by a validated ground truth answer, and the total dataset size reaches 300k. To ensure maximum diversity of queries, semantic deduplication is applied within the collection pipeline, effectively eliminating redundant meanings and preserving a broad range of problem types.

During the preprocessing stage, we perform strict deduplication between the queries and the test set. Specifically, a combination of minihash-based deduplication and semantic deduplication is employed. Only queries passing both checks are retained for subsequent training.

For answer generation, we leverage a strong reasoning language model as the teacher model to sample eight candidate answers for each query. All generated responses undergo automatic filtering, including checks for answer completeness, presence of garbled text, mixing of Chinese and English, and excessive repetition. We then extract the final answers from each generation and verify their correctness against the provided ground truth using a mathematical verifier. Only those answers that are validated as correct are kept in the final data pool.

Finally, for each query with multiple correct answers, we sort the answers by their response length and select the shortest as the optimal form. This ensures that the resulting dataset for long CoT supervised fine-tuning is both high in quality and concise in format.

\subsubsection{Stability Analysis of Thinking Mode Activation}

In our initial design, the  [\textbar BlueThink\textbar] tag was placed at the beginning of the query so as to efficiently trigger the model’s internal reasoning capability—referred to as “thinking mode.” At that stage, when the supervised datasets primarily consisted of in-domain data, this strategy consistently yielded stable activation of thinking mode.
However, as the volume of SFT data increased and the training set expanded to include business-oriented samples characterized by rigid, instruction-driven formats, we observed a notable decline in the reliability of thinking mode activation. Business data, in particular, tended to emphasize precise task execution over flexible or creative reasoning, inherently conflicting with the objectives of the “thinking mode” mechanism.

Further analysis indicated that when the  [\textbar BlueThink\textbar] tag was positioned at the beginning of the query, the presence of strong instruction cues in such business data often overwhelmed or interfered with the expected activation of thinking mode. This effect became more pronounced in out-of-domain scenarios, where data heterogeneity further challenged the robustness of the trigger.
To address this issue, a series of systematic experiments were conducted, exploring alternative positions for the  [\textbar BlueThink\textbar] tag within the query input. Specifically, we repositioned the  [\textbar BlueThink\textbar] tag to follow the query. Empirical results demonstrated that this adjustment led to significantly more stable activation of thinking mode across both in-domain and out-of-domain settings. By placing  [\textbar BlueThink\textbar] at the end of the query, the interference effect of instruction-style prompts was substantially reduced, ensuring the consistent and reliable triggering of internal reasoning capabilities.

\subsubsection{Reinforcement Learning from Human Feedback}
For open-ended question-answering tasks, we implement Reinforcement Learning from Human Feedback (RLHF), a process encompassing preference data collection and reward model training.

\paragraph{Preference Data Construction.}

Our goal is to construct a comprehensive post-training dataset that covers a wide range of task types. To achieve this, we categorize tasks into two main domains: text-only and multimodal. 
We construct a comprehensive training dataset by integrating multiple open-source preference datasets and additional in-house annotations. The multimodal dataset primarily includes the following sources: (1) RLAIF-V-Dataset~\cite{yu2023rlhf, yu2024rlaifv}, (2) LLaVA-OV\footnote{https://huggingface.co/datasets/Fahad-S/LLAVA-OV}, (3)VLfeedback\footnote{https://huggingface.co/datasets/Zhihui/VLFeedback}, (4) LLaVA-RLHF\footnote{https://huggingface.co/datasets/Xiaodong/llava\_rlhf}. For open-source data lacking preference pairs, we employ a high-temperature sampling approach to generate multiple responses using both strong and weak models. These responses are then evaluated by multiple judge models to assess the relative quality of response pairs. We subsequently filter for consensus judgments across the judge models, using the consistently rated pairs as training data for the reward model.

We augment existing datasets with a proprietary in-house dataset focused on Chinese-language data, especial business operational data. The tasks primarily involved carefully text rewriting, summarization and security. we design prompt to judge which response is better through GPT-4o or DeepSeek following the evaluation principles like faithfulness, helpfulness, creativity, Consistency.

\paragraph{Reward Model Training.}

For the safety dimension, we trained a discriminative reward model (RM), while for other general dimensions, we employed the open-sourced reward model Athene-RM-70B~\cite{frickathene}. For text rewriting, document summarization, and other high-volume business scenarios, we trained a generative reward model. This model first evaluates and scores responses across multiple dimensions (such as relevance, accuracy, usefulness, redundancy, etc.), and finally provides an overall score.

For questions with free-form ground-truth answers, we rely on the reward model to determine whether the response matches the expected ground-truth. Additionally, if numerical calculations are involved, we also use the critique reward model to assess whether there are any errors in the mathematical computation process. For multimodal problems with definitive answers, we can also use this text-based reward model to verify the correctness of results. This is because we only need to check whether the final output matches the ground truth and whether the intermediate calculations are accurate—tasks that typically do not require image information.


\subsubsection{Reinforcement Learning with Verifiable Rewards}

\paragraph{Policy Optimization.}

In the reinforcement learning phase, we adopt GRPO (Group Relative Policy Optimization) as the primary optimization algorithm. Unlike the classical PPO (Proximal Policy Optimization), GRPO eliminates the value model and instead estimates the baseline using scores from within the same sample group when computing the advantage.
This design is particularly advantageous in LLM training, as it significantly reduces computational overhead, contributing to its widespread adoption. Furthermore, to enhance training stability, we introduce KL divergence between the current policy and the reference policy into the final loss.

\paragraph{Rule-based reward for reasoning problems.}

In early RL experiments with pure-text-based reasoning models, we demonstrated the effectiveness of rule-based rewards. Building on this finding, we adapted and refined the rule-based reward mechanism for multimodal reasoning RL, with the following components:

\textit{Answer Reward}. For reasoning tasks, we instruct the model to output its reasoning result in a structured format (e.g., /box[]). The output is then compared against the ground-truth answer, and its correctness is verified via predefined rules. Correct and incorrect answers receive rewards of 1 and 0, respectively. For coding tasks, correctness is validated through sandbox execution, where a reward of 1 is granted only if all test cases are passed.

\textit{Format Penalty}. To enforce structured responses, we introduce a format penalty coefficient for violations such as excessive use of /box[] or overly brief reasoning steps. This penalty is multiplied with the answer reward. If penalization is required, the coefficient is set to 0.1; otherwise, it defaults to 1.0.

\textit{Repetition Penalty}. For the repetitive patterns detected in the model's responses, we introduced a repetition penalty coefficient, which is multiplied by the answer reward. Consistent with the format penalty, the repetition penalty coefficient is set to 0.1 if the penalty is to be applied, and remains at 1.0 otherwise. 

The final rule-based reward for multimodal reasoning RL is computed as the product of these three components.

\paragraph{Length penalty and Long2short.}

Large reasoning language models often exhibit what we term an “overthinking” phenomenon when performing complex reasoning tasks. This is characterized by the generation of redundant reasoning steps and unnecessary details, which not only increases response latency but also results in considerable computational waste. To address this challenge, we build upon an innovative length penalty mechanism~\cite{DBLP:journals/corr/abs-2501-12599}, integrated into our Reinforcement Learning (RL) training framework. Our approach aims to reduce the average inference length and improve token efficiency, while maintaining stable training stability.

Our core method relies on a comparative strategy we term “\textit{Group Overlong}.” The principle is to reward or penalize a response based on its length relative to other responses generated for the same query. In each RL step, for each Query x, we sample k different responses, denoted as ($x$, $y_1$), ($x$, $y_2$), $\dots$, ($x$, $y_k$). We compute the minimum and maximum output lengths within the sampled group $L_{min}$  and $L_{max}$ first, and obtain the length difference as follows: $\Delta_L = L_{max}-L_{min}$. To prevent drastic reward fluctuations from destabilizing the training process, we cap $\Delta_L$ as follows: $\Delta_L = max(500, \Delta_L)$.

The length reward, denoted as $R_{len}(i)$, the length penalty is applied only when a response is judged to be correct when $r(x, y_i, y^* )=1$, otherwise, the length reward is zero. The length reward $R_{len}(i)$ is formulated as follows:

\begin{equation}
    R_{len}(i) = \begin{cases}
        \alpha \cdot \lambda_i,\quad\text{if } r(x, y_i, y^*)=1 \\
        0,\;\;\;\;\;\;\;\;\;\; \text{if } r(x, y_i, y^*)=0
    \end{cases}\;,\quad \text{where } \lambda_i = 0.5 - \frac{L_i-L_{min}}{\Delta_L}
\end{equation}


Crucially, the penalty strength $\alpha$ is not fixed but dynamically adjusted throughout training. It decreases following a specified decay schedule based on both the global training step and the absolute length of the current response. This approach prevents the length penalty from diminishing too quickly, which could compromise training stability and convergence in later stages. By default, the base penalty coefficient is set to $\alpha=0.2$, decreasing linearly but bounded below to ensure continued effectiveness.

In summary, our dynamic length penalty mechanism, grounded in the “Group Overlong” strategy, effectively curbs the model’s overthinking behavior and significantly reduces average inference length. Through techniques such as reward clipping and adaptive scaling of $\alpha$, we ensure stable and reliable reinforcement learning. Ultimately, this method enhances inference efficiency and token utilization, achieving a thoughtful balance between training stability and real-world application demands—all while preserving high-quality model outputs.

\begin{figure}
    \centering
    \includegraphics[width=1\linewidth]{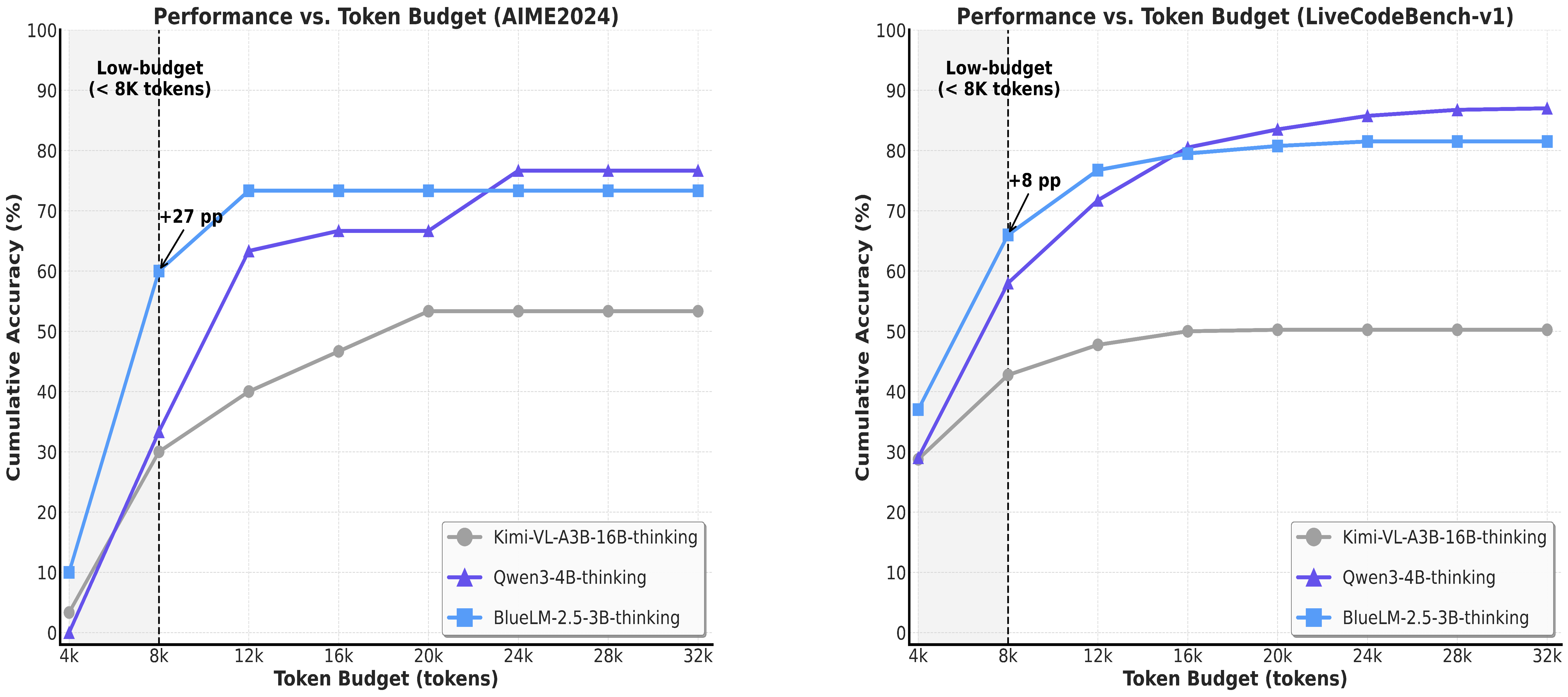}
    \caption{Cumulative accuracy vs. Token Budget on the AIME2024 (left) and LiveCodeBench-v1 (right) benchmarks for Kimi-VL-A3B-16B, Qwen3-4B, and BlueLM-2.5-3B. Our model achieves significant gains in the constrained regime (\textless 8k tokens) while sustaining comparable performance to both baselines at larger context lengths.}
    \label{fig:cumulative_accuracy_maximum_context}
\end{figure}

By integrating the long COT data selection strategy introduced before and the length penalty strategy, our model is transformed into an inference-efficient variant tailored for edge-deployment scenarios. The resulting behavior is shown in Figure~\ref{fig:cumulative_accuracy_maximum_context}, which plots cumulative accuracy as a function of token budget (4K–32K tokens) on two representative benchmarks—AIME2024 (left) and LiveCodeBench-v1 (right)—for three models: Kimi-VL-A3B, Qwen3-4B and the proposed BlueLM-2.5-3B. In the constrained regime (\textless 8K tokens), BlueLM-2.5-3B achieves significant gains, outperforming Qwen3-4B and Kimi-VL-A3B by approximately 27 points on AIME2024, and exceeding Qwen3-4B by about 8 points and Kimi-VL-A3B by about 23 points on LiveCodeBench-v1. For larger context windows (\textgreater
 8K tokens), Our model sustains accuracy within a comparable range to both baselines.

\paragraph{Sampling strategies and other tricks.}

During the preliminary experimental phase prior to formal training, we
explored numerous RL strategies, including but not limited to: (1) Curriculum learning, where progressively easier questions with high model accuracy rates were filtered out as training progressed; (2) Employing a relatively high temperature during the RL sampling phase; (3) Applying entropy bonuses to prevent entropy collapse. However, none of these strategies yielded significant improvements in our experiments, which we will investigate further. Notably, we observed that resetting hyperparameters and resuming training could break stagnation in metric growth, leading to further performance gains. Consequently, we adopted this approach in the final RL phase.

\subsubsection{Hybrid Reinforcement Learning Strategy}

Heterogeneous training data comprise a mixture of math,  code, STEM, instruction following, and mobile service tasks (e.g., text summarization, screen understanding). Each task type employs a dedicated reward mechanism, which may be rule-based or implemented via a separately deployed reward model service accessed through API requests.

We observed that training exclusively on math, code, or STEM data enhances the model's capability in the respective domain without significantly compromising performance in others. Consequently, the proportions of each data type in the mixed training are determined based on the efficacy observed during single-domain training.
Although the reward scales may differ across tasks, advantage normalization is applied to align the magnitude of optimization signals from different data sources. This ensures stable optimization during mixed RL training.

\FloatBarrier

\section{Data Construction}

\subsection{Pure-Text Pre-Training Data }
Since the release of our first-generation large language model in 2023 \cite{bluelm2023}, we have developed a series of models with diverse parameter scales, including 3B, 7B, 13B, and 70B variants. To initialize the LLM parameters of the multimodal model, we first train a text-only model, LLM-base-3B, serving as the foundation for subsequent multimodal training. The training of LLM-base-3B involves a two-stage process: first, a larger 7B model is trained, and then a smaller 3B model is obtained via pruning and distillation. The models are trained using a total of 9.3T tokens, with the 7B model trained on 6.3T tokens and 3B model distilled using an additional 3T tokens. The model trained using this method outperforms a 3B model trained from scratch on 12T tokens.

All training data are sampled from an internal pretraining corpus. In addition to continuous expansion in scale, the dataset emphasizes balance and diversity in content.
The pretraining corpus is composed of a wide range of sources, including:
\begin{itemize}
    \item Publicly available or third-party licensed content from web pages, source code, books, academic papers, general knowledge databases, question-answer pairs, and exam-style problems;
    \item Internally curated industry-specific textual knowledge.
\end{itemize}

Linguistically, the data primarily covers Chinese and English, with supplementary content in over a dozen additional languages.

All raw data undergoes a standardized preprocessing pipeline that includes format conversion, quality filtering, deduplication at multiple granularities, and content classification.

Following the above preprocessing steps, we apply algorithmic quality assessment to evaluate each sample and select high-quality data. An automated algorithm is then used to determine the optimal sampling ratio across data sources to maximize model performance.

\subsection{Multimodal Pre-Training Data}

The volume of multimodal pre-training data has been expanded significantly, increasing from 1T tokens in BlueLM-V-3B \cite{lu2025bluelm} to 4T tokens in the current version. Of this, 1.6T tokens are pure text data and 2.4T tokens consist of image-text pairs. Notably, pure text now accounts for 40\% of the training data, substantially higher than the approximately 5\% in BlueLM-V-3B.

The training data spans a wide range of categories, including:

\begin{itemize}
    \item Caption Data: Image captioning data with detailed textual descriptions.
    \item OCR Data: Optical character recognition datasets containing images with embedded text and corresponding transcriptions.
    \item VQA Data: Visual question answering datasets with question-answer pairs grounded in visual inputs.
    \item GUI Data: Multimodal data related to graphical user interfaces and virtual agent interactions.
    \item Pure-Text Data: High-quality monomodal text data integrated to preserve and enhance language understanding capabilities.
    \item Other Data: A mix of knowledge-intensive datasets, interleaved text-vision content, and proprietary in-house sources.
\end{itemize}

This expanded and diversified dataset composition aims to improve the model's performance across a wide range of multimodal reasoning, perception, and generation tasks.

\subsubsection{Image Caption Data}
The image caption dataset combines extensive open-source Chinese and English caption data with substantial high-quality in-house image-text pairs, ensuring broad coverage and diversity. A rigorous multi-stage process involving deduplication, quality filtering, and resolution variation maintains high data quality and strong image-text alignment. To address class imbalance, especially in niche scenarios, the dataset leverages a well-structured internal category system and uses open-source captioning models to generate captions for existing category images, effectively supplementing scarce data. This approach results in a balanced, diverse, and high-quality caption dataset that significantly enhances multimodal model generalization and performance across common and rare categories.

\subsubsection{OCR Data}
The OCR dataset is composed of both real annotated data and synthetic data to ensure comprehensive coverage and robustness. The real data, collected from both open-source and in-house sources, spans complex and diverse scenarios including office, study, and daily life environments with varied shooting conditions. Complementing this, the synthetic data is generated at multiple granularities—such as lines, paragraphs, pages, and vertical text layouts—using advanced 3D rendering techniques that realistically simulate distortions, shadows, wrinkles, and other natural artifacts. The dataset supports multiple annotation formats like Markdown, HTML, and grounding, enabling a wide range of downstream tasks including text extraction, formula extraction, document reconstruction, and table recognition. Additionally, it covers a broad linguistic spectrum, incorporating not only Chinese and English but also minority languages, rare Chinese characters, and pinyin, thus providing rich multilingual and multi-script support for robust OCR performance across diverse real-world scenarios.

\subsubsection{GUI Data}
GUI data includes diverse, high-quality GUI grounding, referring, and trajectory data. To enhance the model’s screen comprehension capabilities, we assemble a dataset comprised of open-source and in-house data from various platforms, including web pages, mobile devices, and computer desktops. This dataset covers both widget caption and instruction grounding, facilitating a deeper understanding of the descriptions and functionalities of GUI components. Additionally, we have automated the construction of extensive Chinese-language GUI data to strengthen the model's grasp of Chinese interfaces. For GUI decision-making, we collect a large amount of multi-step annotated trajectories from open sources, and unify the trajectory operations into a standardized action space.

\subsubsection{Pure Text and Other Data}
The pure text data used here is sampled from the previously mentioned pure-text pretraining corpus.

In addition, the pretraining corpus also includes multimodal knowledge data in image-text format, grounding data, interleaved image-text data, and other proprietary internal data combining visual and textual content.

\subsection{Reasoning-Enhanced Data}
To substantially enhance the reasoning capabilities of our small-sized model, we have curated a large-scale corpus specifically designed for reasoning tasks. Our training data consists of two main components: approximately 950 billion tokens of pre-training data and approximately 450 billion tokens of synthetic data. For the pre-training data, we focus on a carefully selected subset from general pre-training corpora that is highly relevant to STEM, programming, and logical reasoning. This ensures both \textbf{broad knowledge coverage} and \textbf{data diversity} during the reasoning enhancement phase.

For the synthetic data, we design a two-stage pipeline:
(1) question collection and augmentation, and
(2) answer generation.
In the first stage, we gather a large number of raw questions from multiple sources, including direct collection as well as extraction from pre-training corpora or PDF documents~\cite{numina_math_datasets}. Following rigorous deduplication and contamination filtering, we utilize high-performing large language models, such as the latest releases from the Qwen and DeepSeek series, to augment the original questions~\cite{yu2024metamathbootstrapmathematicalquestions,tang2024mathscalescalinginstructiontuning}, thereby enhancing the dataset with higher diversity and effectively increasing its overall scale.
In the second stage of synthetic data construction, we generate high-quality reasoning paths using teacher models specialized for two reasoning styles: long-chain-of-thought and short-chain-of-thought. Specifically, we perform diverse sampling to obtain multiple reasoning trajectories per question.

To ensure answer correctness, we apply a multi-pronged quality control strategy combining rule-based text sanitization, rejection sampling~\cite{yuan2023scalingrelationshiplearningmathematical}, and majority voting~\cite{wang2023selfconsistencyimproveschainthought}. For questions with deterministic answers, we use rejection sampling to discard incorrect generations. In cases where ground truth is unavailable, we employ majority voting across multiple model outputs to select the most plausible response.
Regarding multimodal data, we utilized approximately 30 billion of synthetic reasoning data, encompassing mathematical problem solving and domain-specific VQA. Modeled after the text data production pipeline, we employed powerful large-scale multimodal reasoning models combined with curated reasoning prompts to extract detailed reasoning trajectories. The quality of the reasoning data was further enhanced through rule-based checking, rejection sampling, and vote-based techniques.


\subsection{Long-Context Data} 
To enable the model to activate long-context capabilities for pure-text and multimodal inputs, the long-context data in our framework includes extended textual sequences and multimodal data, such as interleaved text-image corpora (image QA, visual descriptions, OCR results). To enhance long-form reasoning, we supplemented native long-text data with high-quality long-range reasoning datasets, ensuring robust contextual understanding across modalities.

\subsection{Post-Training Data}

\subsubsection{Instruction Data}

To enhance the model's conversational and instruction-following capabilities, we constructed a high-quality instruction dataset through a rigorous multi-stage process. Initially, we collected a substantial volume of queries and images from both open-source communities and internal business sources. Subsequent labeling and classification revealed significant redundancy in the query data. To ensure diversity, we systematically eliminated repetitive single-instruction queries and reconstructed new queries along defined competency dimensions.

For answer generation, we leveraged multiple state-of-the-art models to synthesize responses, followed by a comprehensive quality control pipeline, including rule-based filtering of low-quality answers, model-based scoring mechanisms, majority voting systems, and manual verification by human experts.

The final dataset was carefully balanced to reflect real-world application scenarios, maintaining an approximately 1:1 ratio between Chinese and English language content, as well as between textual and multimodal instruction samples.

This methodology guarantees exceptional diversity, quality, and balance in three critical aspects: content coverage across domains, linguistic representation, modality distribution.

To enhance the performance of large language models in reasoning tasks of varying complexity, we have developed an innovative “thinking mode” switching mechanism. This approach leverages a dedicated control token,  [\textbar BlueThink\textbar], which determines whether the model engages its “long-thinking” mode.

During supervised fine-tuning, we curated a large dataset annotated with special tags to train the model’s capacity for complex reasoning. To enable mode switching during inference, we introduce the  [\textbar BlueThink\textbar] token to explicitly signal the activation of long-thinking mode. During training, samples requiring in-depth reasoning are prepended with  [\textbar BlueThink\textbar], allowing the model to learn to adjust its reasoning process according to the presence of this token. Conversely, content following the tag is also incorporated as short-thinking examples in a mixed training regime but is presented without the  [\textbar BlueThink\textbar] token.

When users encounter logically challenging problems that require multi-step deduction, they can append  [\textbar BlueThink\textbar] to their query. Upon detecting this token, the model switches to “long-thinking” mode, generating comprehensive, structured reasoning (enclosed in \texttt{<think>} ... \texttt{</think>} tags) that simulates deep human analysis and significantly improves response accuracy and 
reliability. For simpler queries with straightforward answers, users can omit the token, in which case the model defaults to a “short-thinking” mode—skipping unnecessary intermediate steps for faster, more efficient responses.

To evaluate the robustness of this mechanism, we conducted extensive real-world testing. The results demonstrate that the  [\textbar BlueThink\textbar] switch has an exceptionally low failure rate—less than 1 PPM (fewer than 1 failure per million uses)—highlighting the outstanding stability and reliability of this control mechanism in practical applications.

\subsubsection{RL Data} 
The RL training queries encompass the following categories: math, code, STEM, instruction following, and mobile service tasks. Math data was collected from open-source datasets DAPO~\cite{yu2025dapoopensourcellmreinforcement}, DeepScaler~\cite{deepscaler2025}, ORZ~\cite{hu2025openreasonerzeroopensourceapproach}, Numina~\cite{numina_math_datasets}, deduplicated, and subjected to small-scale RL experiments, queries were retained only if the average response length and mean accuracy across all sampled responses met specific thresholds. Code data was sourced from the OpenR1~\cite{openr1} project. STEM data originated from the SCP-116K Dataset~\cite{lu2025scp116khighqualityproblemsolutiondataset}. We retained queries containing an extract solution and the solution was validated against the R1 response. Instruction following data was generated using DeepSeekV3. We composed queries adhering to IFEval's format specifications, prompted DeepSeekV3 to generate corresponding questions, and filtered out queries where DeepSeekV3 failed to produce correct answers. The final dataset comprises 50k math queries, 30k code queries, 15k STEM queries, 30k instruction following queries and 20k mobile service task related queries.


\subsection{Pre-Training Data Efficiency}



As illustrated in Table~\ref{tab:training_data_compare}, BlueLM-2.5-3B uses 23\% less total pre-training data (including both LLM and MLLM pre-training data) compared to similar scale models such as Qwen2.5-VL-3B. This reduction is primarily due to the efficient control of the text model's pre-training volume. The pure-text LLM model used to initialize BlueLM-2.5-3B was trained on only 25\% of the data used by Qwen3-4B, and just 48\% of that used for initializing Qwen2.5-VL-3B via Qwen2.5.

Another key distinction lies in the volume of multimodal pretraining data: BlueLM-2.5-3B incorporates substantially more than the other models. This is largely because the reasoning capabilities of the text model were enhanced in later training stages and integrated into the multimodal phase. Considering the upsampling of key samples, a total of 3.3T tokens of reasoning-enhanced data was used across these stages—forming a solid foundation for the model’s strong reasoning performance despite its small size.

\begin{table}[ht]
    \centering
    \caption{Training Data Amount Comparison}
    \renewcommand{\arraystretch}{1.2} 
    \resizebox{0.9\textwidth}{!}{
    \begin{tabular}{l C{5cm} C{5cm} C{5cm}}
        \toprule
        \textbf{Model} & \textbf{LLM Pre-training} & \textbf{MLLM Pre-training} & \textbf{Total Pre-training} \\
        \midrule
        Qwen3-4B & 36T & - & 36T \\
        Qwen2.5-VL-3B & 18T & 4.1T & 22.1T \\
        Kimi-VL-A3B-16B & 5.2T & 4.4T & 9.6T \\
        \midrule
        \textbf{BlueLM-2.5-3B} & 9.3T & 7.8T & 17.1T \\
        \bottomrule
    \end{tabular}
    }
    \label{tab:training_data_compare}
\end{table}

\subsection{Data Pipeline}

\begin{figure}
    \centering
    \includegraphics[width=0.9\linewidth]{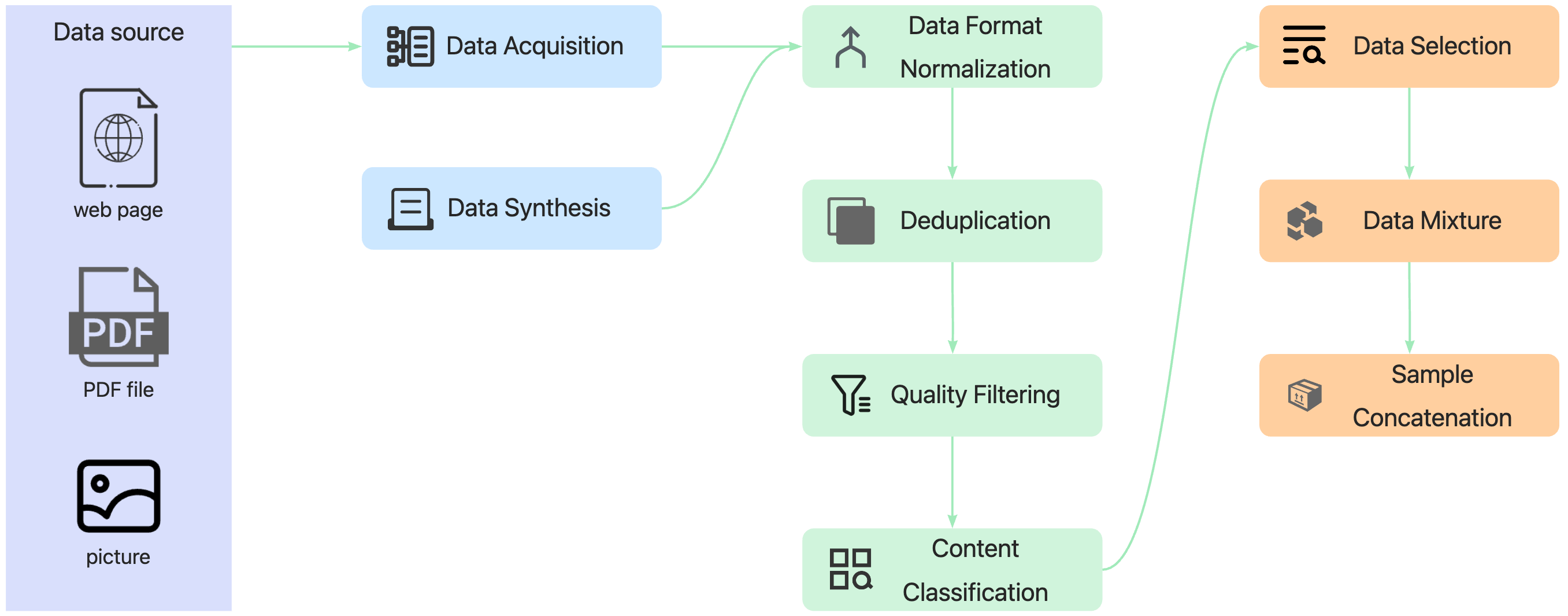}
    \caption{Data Pipeline of BlueLM-2.5-3B.}
    \label{fig:data pipeline}
\end{figure}

To support efficient data curation and sample construction, we developed a comprehensive data pipeline that spans the full lifecycle of data processing. As shown in Figure~\ref{fig:data pipeline}, the pipeline includes modules for data acquisition, synthetic data generation, format conversion, quality filtering, deduplication, content classification, data selection, data mixture, and training sample construction.

\subsubsection{Data Synthesis} 
To enhance dataset coverage, particularly in data-scarce domains, we employ open-source models, including the latest versions of Qwen and DeepSeek, to generate synthetic data with controlled authenticity and consistency. Our data synthesis pipeline consists of two distinct phases: model selection performed on single-node, and batch production implemented through distributed approaches.

Both methodologies utilize the widely-adopted, high-performance vLLM framework as the inference engine. The system exposes model configurations and sampling parameters through a configurable interface, enabling rapid adaptation to diverse models with minimal modifications.

For the distributed implementation, we established an efficient synthesis architecture featuring:
1. A message queue (implemented with Pulsar) connecting service nodes to avoid rebalancing issues during node scaling
2. Kubernetes-based dynamic node orchestration for elastic resource allocation

Notably, we observed significant diurnal variation in online inference workloads, with peak demand during daytime and underutilized GPU capacity at night. To optimize resource efficiency, we developed an adaptive scheduling mechanism that: leverages idle GPU capacity during off-peak periods; increases synthetic data throughput by 2.3× (empirical measurement); maintains baseline service-level agreements for production inference tasks.

This dual-phase synthesis framework demonstrates three key advantages:
a) Scalability: Processes 12M tokens/hour/node at peak capacity.
b) Resource efficiency: Achieves 78\% GPU utilization during traditional idle periods.
c) Adaptability: Supports 15+ foundation model variants without architectural changes.

\subsubsection{Format Conversion}

Format conversion refers to the process of extracting relevant textual content from diverse source formats—such as HTML, PDF, and tables—and transforming it into a standardized text format suitable for downstream processing and model training.

\subsubsection{Data Filtering} 
The quality of training data serves as the decisive factor in determining the performance of LLMs and MLLMs. Pure text quality filtering is implemented using a combination of rule-based heuristics and model-based scorers, which evaluate the data across multiple quality dimensions.

For vision-text data, we implemented a two-stage filtering pipeline designed to improve data integrity throughout the entire process:

Stage 1 Domain-specific filtering and content moderation. 
We first employed open-source vision models to generate standardized textual descriptions of visual data. These descriptions were then precisely categorized into 31 predefined domains (e.g., knowledge, programming) using a pretrained text classification model, which guided data balancing. Concurrently, we identified and filtered out meaningless, low-quality, harmful, and non-wholesome content, substantially improving the overall quality of the training data.

Stage 2 Heuristic-based quality control. 
Building upon the results from stage 1, we further applied predefined heuristic rules to eliminate problematic entries, including: anomalously repetitive data, truncated model responses, and ethically harmful content.

This secondary filtering stage prevents the model from learning erroneous patterns, thereby enhancing generalization capability.

\subsubsection{Deduplication}

Our deduplication strategy combines N-Gram-based filtering and MinHash-based similarity detection at multiple granularities. It is applied at multiple levels, including: 

\begin{itemize}
    \item Intra-training set deduplication, especially across samples from different data sources.
    \item Cross-set deduplication, ensuring no overlap between the training set and both the validation set and evaluation benchmarks, a process commonly referred to as \textit{decontamination}.
\end{itemize}

This comprehensive approach minimizes bench data leakage and ensures the reliability of downstream evaluation.

\subsubsection{Content Classification}

Drawing on existing classification taxonomies and informed by data clustering results, we defined a comprehensive set of several hundred categories. A multi-class classifier was trained using a lightweight model (BERT) to accurately assign category labels. This classifier is subsequently used to perform fine-grained content classification on each individual data sample.

\subsubsection{Data Selection}

The objective of data selection is to identify high-quality samples that contribute to improved model performance. Common approaches include rule-based filtering based on expert heuristics and model-based methods that leverage trained classifiers or scoring functions. In collaboration with HKUST, we propose a novel data selection method called \textit{PreSelect} \cite{shum2025predictive}. This method introduces the concept of \textit{Predictive Strength}, along with a corresponding computation framework. Predictive Strength quantifies a sample's contribution to a specific model capability by analyzing the loss ordering across different models, capturing how consistently a sample supports or challenges model behavior. Using this metric, we identify effective samples for a given capability and use them to train a fastText-based classifier, which is then applied to the entire pretraining corpus for large-scale filtering. By leveraging this approach, we are able to selectively extract subsets of data from the existing pre-training data corpus that exhibit stronger alignment with the targeted model capabilities, thereby enhancing the efficiency and effectiveness of the pretraining process.

\textbf{Data Augmentation.} 
As the scale of the pretraining corpus increases, a key challenge emerges: certain data sources or categories—while critical for specific capabilities or overall model performance—are limited in volume and cannot be scaled up directly. A common approach is oversampling these data types; however, this method has inherent limitations, as the number of effective repetitions is bounded, and shortages may persist even after reaching the duplication threshold. To address this, in addition to traditional oversampling, we adopt a method known as \textit{DSIR} \cite{xie2023data}. DSIR enables us to retrieve data similar to the target category from a broader pool of web-scale data. Using this approach, we have effectively expanded long-tail categories such as encyclopedic content, academic papers, and world knowledge, thereby enhancing coverage in these high-value but low-resource domains.

\subsubsection{Data Mixture}

The pretraining corpus comprises diverse sources, modalities, and data types, and determining the optimal proportion of each data component is a critical factor that directly affects model performance. In the context of multimodal model training, there is currently no established consensus on the ideal ratio between pure-text and image-text data. Likewise, the proportions of various data subcategories are often set based on heuristics or limited ablation studies. To address this, we applied the \textit{RegMix} \cite{liu2024regmix} method for automated mixture optimization in the pretraining of our text-based models, demonstrating superior performance compared to manually tuned, prior-based strategies. Building on this success, we extended RegMix to support automatic data ratio optimization in the multimodal setting as well.

\textbf{Balancing Between Pure Text and Multimodal Data.} 
A key challenge in multimodal model training arises when initializing the LLM parameters with a pretrained text model—such initialization often leads to catastrophic forgetting of text understanding capabilities during subsequent multimodal training. Our previous experimental results show that during the image-text joint pretraining phase, techniques such as reducing the learning rate or freezing portions of the LLM parameters can effectively mitigate the degradation of textual capabilities. However, these approaches come at a significant cost—namely, a notable decline in multimodal understanding performance. In contrast, increasing the proportion of pure-text data during training has proven to be an effective strategy that preserves multimodal learning performance while simultaneously alleviating the forgetting of textual capabilities.

\subsubsection{Sample Concatenation} 
To enhance training efficiency and implement a diversified sample concatenation strategy with load balancing, we preprocess data into fixed-length segments prior to training. For this purpose, we developed a high-performance data concatenation framework with the following core innovations:

\begin{enumerate}
    \item Similarity-Based Data Aggregation. 
Data samples are clustered according to their vector similarity before concatenation, ensuring semantic coherence within each concatenated unit and effectively preserving data consistency.

    \item Semantic-Preserving Greedy Concatenation.  
We employ a greedy algorithm optimized to: minimize fragmentation of complete data samples; Maintain semantic integrity as the highest priority; precisely control output length to approach target size; balance multimodal (text-image) content ratios

    \item Large-Scale Distributed Execution. 
The system leverages dynamic scheduling capabilities of our massive computing cluster (10,000+ core resource pool) to achieve: sub-second task response times; linear scalability with cluster size; fault tolerance during distributed processing

\end{enumerate}

To ensure data quality and prevent evaluation contamination, our preprocessing pipeline performs rigorous deduplication between training, validation, and test sets using both text embeddings and image features. We implemented a centralized data management platform providing integrated capabilities for data mixing and blending, visual analytics, dataset shuffling, Batch packaging.

\FloatBarrier

\section{Training Infrastructure}

\subsection{Training Cluster}
\subsubsection{Computing, Networking, and Storage}
We have built an on-premises GPU cluster of thousands of high-performance GPUs. These GPU servers are interconnected via a 4×200 Gb/s InfiniBand (IB) network, adopting a fat-tree and rail-optimized topology to achieve high-efficiency communication, with a speedup ratio exceeding 95\% in thousand-GPU training. For storage, we developed the XuanYuan File Storage System, which enhances I/O performance through optimizations such as tiered read caching, write caching, and metadata caching.

\subsubsection{Stability}
Stability remains a critical challenge in large-scale training~\cite{jiang2024megascalescalinglargelanguage, grattafiori2024llama3herdmodels}. Through systematic optimizations, we have achieved a 99\% effective training uptime.

\textbf{Reducing Hardware Failure Rates.}
1. Early-stage challenges: Initial hardware failure rates were high. By targeting recurring issues (e.g., ECC errors, NaN computations), we reduced the daily failure rate from 2\% to 0.1\%.
2. Proactive exclusion: Historical data indicates that machines with prior failures exhibit a higher likelihood of recurring faults, even after repair. Thus, we proactively exclude such machines before large-scale training.

\textbf{Optimizing Fault Handling Procedures.}
1. Pre-training: Comprehensive diagnostic tools are used for hardware defects, network instability, and performance degradation (e.g., faulty/slow nodes), ensuring only healthy nodes are chosen for training.
2. During training: An automated monitoring system tracks server health, network status, and job performance. Lightweight diagnostic tools enable real-time anomaly detection, isolation, and task recovery, minimizing downtime.
3. Post-training: Newly encountered failures are analyzed to enhance diagnostic capabilities, enabling faster identification and resolution of similar issues in subsequent runs.

\FloatBarrier
\subsection{Pre-Training Framework}

We developed an in-house training framework based on Megatron-LM~\cite{shoeybi2020megatronlmtrainingmultibillionparameter} for training LLM and VLM. Our optimizations focus on four key aspects: training efficiency, scalability, stability, and observability.

\subsubsection{Efficiency}

\begin{figure}[H]
    \centering
    \includegraphics[width=0.9\linewidth]{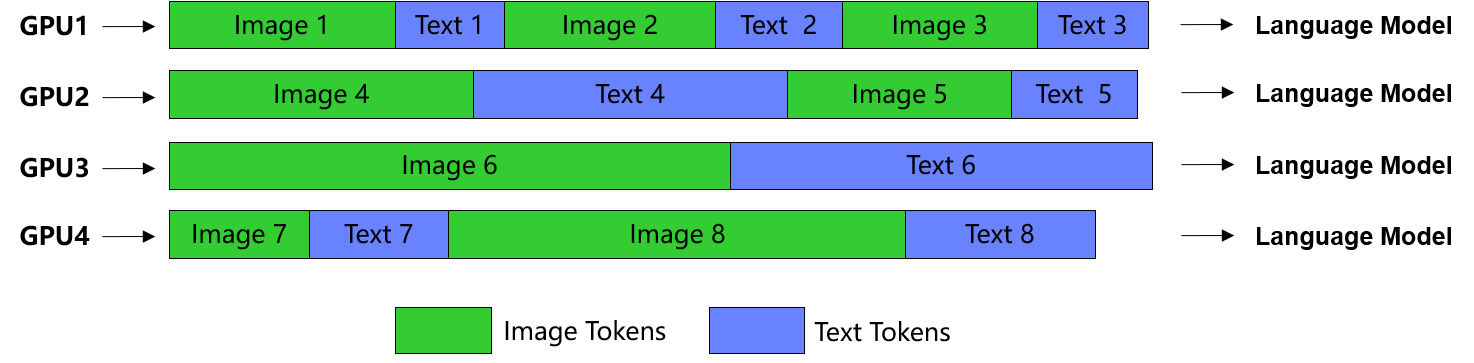}
    \caption{Multi-Sample Concatenation. Multiple image–text pairs are interleaved into a single composite sample per GPU, balancing image and text token counts to ensure a uniform computational load and maximize GPU utilization.}
\label{fig:sample concatenation}
\end{figure}

\textbf{Data Concatenation.}
Image-text pairs often vary significantly in image resolution and text length. Processing a single pair per GPU leads to imbalanced computational load and resource wastage. To address this, as shown in Figure~\ref{fig:sample concatenation}, we concatenate multiple image-text pairs into composite samples, balancing the number of image tokens, text tokens, and total tokens per sample to maximize GPU utilization.

\begin{figure}[t]
    \centering
    \includegraphics[width=1.0\linewidth]{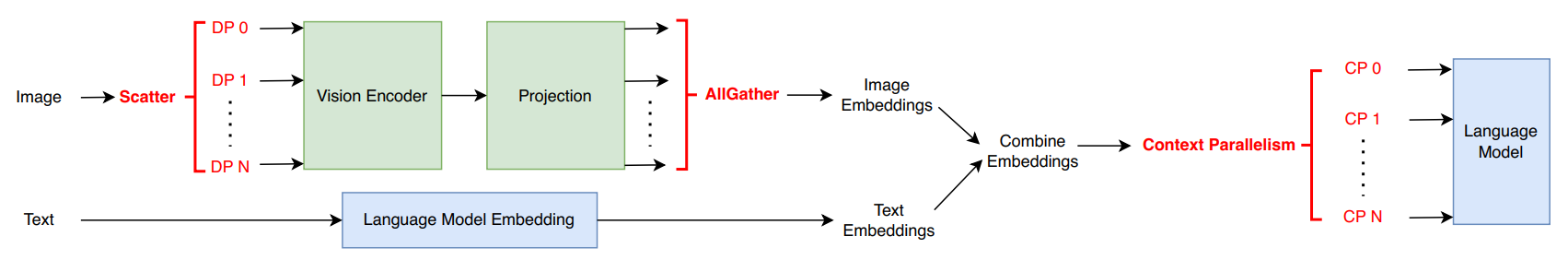}
    \caption{Long-Context Training. Combined image and text embeddings are split into N equal-length subsequences and processed in parallel on separate GPUs (context parallelism) to enable long-sequence training. Crucially, by applying data parallelism to the original image inputs within each context-parallel group—having each GPU concurrently perform vision encoding and projection—we achieve a 1.66× improvement in overall training throughput. Losses from all subsequences are then aggregated for end-to-end gradient computation.}
\label{fig:long sequence training}
\end{figure}

\textbf{32K Long-Context Training.}
To enable 32K long-sequence training, we employ context parallelism~\cite{liu2023ringattentionblockwisetransformers}, splitting input embeddings into equal-length subsequences (Figure~\ref{fig:long sequence training}). Each GPU processes a subsequence in parallel during forward and backward passes, with losses aggregated to compute the final gradient update, enabling efficient long-context training. Building on this, we innovatively applied data parallelism to image data within each CP group, allowing multiple GPUs in the group to concurrently load and process images, thereby further enhancing training performance.

\textbf{Linear Scaling with Ultra-Large Batch Sizes.}
To shorten training time, we expanded the cluster from 1,024 to 2,048 GPUs and proportionally increased the batch size to 16,384. At the same time, we resolved the instability inherent to ultra-large batches and ultimately achieved a 96.8\% speedup ratio.

\FloatBarrier

\subsubsection{Scalability}

Our framework ensures flexible component integration and cross-framework compatibility, significantly improving experiment efficiency and maintainability.

\textbf{Modular Architecture.}
The vision encoder, adapter layer, and language model are decoupled into independent modules with standardized interfaces. Model components and data processing strategies are plug-and-play, improving experiment efficiency and reducing maintenance overhead.

\textbf{Configuration-Driven Workflow.}
A single configuration file specifies different ViT architectures, adapter layers, and preprocessing/embedding strategies, enabling automatic loading and initialization.

\textbf{Compatibility with Distributed Frameworks.}
The framework seamlessly integrates with mainstream distributed training frameworks such as Megatron-LM~\cite{shoeybi2020megatronlmtrainingmultibillionparameter} and DeepSpeed\footnote{https://github.com/deepspeedai/DeepSpeed}, ensuring code reusability across different cluster environments.

\subsubsection{Stability}
We enhanced fault tolerance by refining exception handling and integrating with the training platform to monitor key metrics. The system triggers alerts for training hangs, performance fluctuations, or loss spikes, with automatic recovery for resolvable failures. Ultimately, we completed a 4.8-trillion-token pretraining task on 2,048 GPUs in 139 hours without interruptions.

\subsubsection{Observability}
To diagnose training instability, we implemented fine-grained monitoring inspired by~\cite{molybog2023theoryadaminstabilitylargescale}, tracking: L2 norms of parameters and gradients; Distributions, norms, and angles of $u$ and $r$ in AdamW optimization; Mean and variance of layer-wise residuals.
These metrics enable rapid anomaly detection and root-cause analysis.

Additionally, we integrated online evaluation, periodically pausing training to assess model performance on benchmark datasets, ensuring real-time progress tracking.

\subsection{RL Infrastructure}

We conducted customized development and performance optimization based on the open-source training framework veRL~\cite{Sheng_2025}, adapting it to the architecture and training methodology of the BlueLM-2.5 large model to enhance training efficiency.

\subsubsection{Performance Optimization}

\textbf{One-Step Async RL.}
The original veRL framework employed a hybrid engine that executed generation and training phases sequentially, leading to inflexible resource scheduling and suboptimal GPU utilization. Building upon An Async Pipelined Version of veRL\footnote{https://github.com/agentica-project/verl-pipeline}, we implemented the following improvements to the training pipeline:
Decoupled Execution: The inference nodes handle generation and reward computation, while the training nodes execute reference model inference, advantage calculation, and training. By adjusting the resource allocation ratio between inference and training nodes, we balanced the execution time of both phases.
Optimized Communication: We replaced the Ray serialization-based communication framework with NCCL, significantly reducing the communication overhead for vLLM parameter updates.
These optimizations achieved an overall 1.32× performance improvement in multimodal training scenarios.

\textbf{Inference Engine Load Balancing.}
To address the bubble time and load imbalance caused by variable sequence lengths in RL generation tasks, we redesigned the ChatScheduler based on veRL’s asynchronous vLLM mode. This enables dynamic and continuous request scheduling, maximizing inference throughput~\cite{fu2025areallargescaleasynchronousreinforcement}.

\subsubsection{Remote Deployment of the Reward Model} 
In practical reinforcement learning training, we fine-tune and deploy multiple reward models tailored to different tasks (e.g., code generation, mathematical reasoning, security policies). The stability, elasticity, and latency of their online deployment directly impact training efficiency and convergence. Therefore, our RM service deployment strategy focuses on three key aspects:

\textbf{Automated Deployment.} Upon passing validation tests, the trained RM model is automatically deployed across multiple instances.

\textbf{Elastic Scaling.} Each RM instance is containerized, enabling dynamic scaling based on demand.

\textbf{Service Registration.} RM instances register with a naming service for automatic discovery and invocation via identifiers.

\FloatBarrier

\section{Evaluation} 

We evaluate and analyze our proposed BlueLM-2.5-3B model on more than 20 benchmark datasets, with detailed descriptions of these benchmarks provided in Table~\ref{tab:multimodal_intro} and Table~\ref{tab:pure_text_intro} in Appendix~\ref{app:bench_list}. For our model in the thinking mode and non-thinking mode, we utilize a sampling temperature of 0.6, a top-p value of 0.95, and a top-k value of 20, consistent with the thinking mode configuration used in Qwen3-4B~\cite{yang2025qwen3}. For the thinking mode, the max output length is set to 32768, while for the non-thinking mode, it is set to 4096. All benchmark datasets are assessed using OpenCompass\footnote{https://opencompass.org.cn/home}.

\subsection{Evaluation Results in the Thinking Mode}

For the comparison of model performance in the thinking mode, we select the following baseline models: leading reasoning models of comparable scale (Qwen3-4B-thinking~\cite{yang2025qwen3}), larger-scale influential reasoning models (Kimi-VL-A3B-16B-thinking-2506~\cite{kimiteam2025kimivltechnicalreport}), and leading closed-source reasoning models (Gemini-2.5-pro\footnote{https://deepmind.google/models/gemini/pro/}, OpenAI-o4-mini\footnote{https://openai.com/index/introducing-o3-and-o4-mini/}).
To further demonstrate model’s impressive reasoning capabilities on benchmarks, the larger-scale non-thinking general model Qwen2.5-VL-72B ~\cite{bai2025qwen25vltechnicalreport} is included for comparative analysis. Additionally, for long-chain reasoning evaluation tasks, we develop and deploy an assessment model that achieves a 6-fold improvement in evaluation speed while reducing API costs by 50\%, with scoring consistency rates comparable to or exceeding that of OpenAI-GPT-4.1\footnote{https://openai.com/index/gpt-4-1/}.


\subsubsection{Evaluation Results of Multimodal Benchmarks} 

As shown in Table~\ref{tab:multimodal-benchmark}, when compared to larger-scale models such as Kimi-VL-A3B-16B-thinking-2506, the performance gap on most benchmarks remains within 5\%.
Thanks to the reasoning-enhanced data, our model even achieves superior performance compared to Qwen2.5-VL-72B on reasoning benchmarks MathVision and MathVista, although a performance gap remains on other benchmarks.

\begin{table}[ht]
    \centering
    \caption{Comparison of BlueLM-2.5-3B-thinking and other baselines of Multimodal Benchmarks. The highest and second-best scores among open-source models are shown in \textbf{bold} and \underline{underline}, respectively.}
    \renewcommand{\arraystretch}{1.2} 
    \resizebox{0.98\textwidth}{!}{
    \begin{tabular}{l C{3.0cm} | C{3.5cm} C{2.8cm} | C{2.8cm} C{2.8cm} }
        \toprule
        \textbf{}& \textbf{BlueLM-2.5-\newline3B-thinking} & \textbf{Kimi-VL-A3B-16B-thinking-2506} & \textbf{Qwen2.5-VL-72B} & \textbf{Gemini-2.5-\newline pro} & \textbf{OpenAI-o4-mini} \\
        \midrule
        Params & 2.9B & A3B-16B & 72B & - & - \\
        \midrule
        MMMU & 51.3 & \underline{64.0} & \textbf{68.2} & 74.7 & 81.6 \\ 
        MMBench-V1.1 & 78.3 & \underline{84.4} & \textbf{87.8} & 88.3 & 90.0 \\
        MMstar & 66.3  & \underline{70.4} & \textbf{70.5} & 73.6 & 76.0 \\
        MM-vet & 65.1  & \textbf{78.1} & \underline{76.9} & 83.3 & 70.5 \\
        OCRBench & 84.0 & \underline{86.9} & \textbf{88.2} & 86.2 & 78.4 \\
        AI2D & 82.6  & \underline{85.3} & \textbf{88.5} & 89.5 & 84.3 \\
        MIA-bench & \underline{82.6} & 82.5 & \textbf{87.8} & 91.0 & 90.3 \\
        MathVista & \underline{78.4} & \textbf{80.1} & 74.2 & 80.9 & 84.3 \\
        MathVision & \underline{47.7}  & \textbf{56.9} & 38.1 & 69.1 & 71.0 \\
        HallusionBench & \underline{57.3}  & \textbf{60.5} & 54.6 & 64.1 & 63.0 \\
        \bottomrule
    \end{tabular}
    }
    \label{tab:multimodal-benchmark}
\end{table}

\FloatBarrier

\subsubsection{Evaluation Results of Pure-Text Benchmarks}

\begin{table}[ht]
    \centering
    \caption{Comparison of BlueLM-2.5-3B-thinking and other baselines of Text Benchmarks. The highest and second-best scores among open-source models are shown in \textbf{bold} and \underline{underline}, respectively.}
    \renewcommand{\arraystretch}{1.2} 
    \resizebox{0.98\textwidth}{!}{
    \begin{tabular}{l C{2.3cm} C{2.3cm} | C{3.5cm} C{2.2cm} | C{2.2cm} C{2.2cm} }
        \toprule
        \textbf{} & \textbf{BlueLM-2.5-3B-thinking} & \textbf{Qwen3-4B-thinking} & \textbf{Kimi-VL-A3B-16B-thinking-2506} & \textbf{Qwen2.5-VL-72B} & \textbf{Gemini-2.5-pro} & \textbf{OpenAI-o4-mini} \\
        \midrule
        Params & 2.9B  & 4.0B & A3B-16B & 72B & - & - \\
        \midrule
        MMLU-pro & 66.7 & \underline{67.9} & 66.4 & \textbf{71.2} & 76.9 & 81.5 \\
        GPQA-Diamond & \textbf{59.6} & \underline{55.6} & 48.0 & 47.5 & 73.7 & 77.8 \\
        Math-500 & \underline{92.8} & \textbf{95.8} & 91.2 & 83.0 & 98.0 & 92.6 \\
        GSM8K & \textbf{95.5} & 94.7 & 92.6 & \underline{95.3} & 95.2 & 95.5 \\
        BBH & \textbf{89.3} & \underline{85.4} & 83.5 & 82.2 & 92.9 & 89.5 \\
        AIME24 & \underline{73.3} & \textbf{76.7} & 53.3 & 20.0 & 90.0 & 86.7 \\
        AIME25 & \textbf{66.7} & \underline{65.6} & 33.3 & 13.0 & 83.0 & 92.7 \\
        IF-Eval & 80.2 & \underline{81.7} & 66.5 & \textbf{86.3} & 88.7 & 90.6 \\
        Humaneval & \underline{94.5}  & \textbf{95.1} & 82.3 & 87.8 & 97.6 & 98.2 \\
        LiveCodeBench-v1 & \underline{81.5} & \textbf{86.8} & 50.3 & 55.8 & 91.8 & 91.8 \\
        LiveCodeBench-v5 & \textbf{54.2} & \textbf{54.2} & 31.3 & 26.5 & 69.9 & 74.7 \\
        \bottomrule
    \end{tabular}
    }
    \label{tab:text-benchmark}
\end{table}

As shown in Table~\ref{tab:text-benchmark}, in comparison with Qwen3-4B-thinking, our model performs competitively across most evaluation tasks, and showing leading results on 4 out 11 tasks.
Against larger-scale models such as Kimi-VL-A3B-16B-thinking-2506, our model outperforms across the board, highlighting its strong language modeling capabilities. On reasoning benchmarks such as Math-500, GSM8K and AIME, our model also outperforms Qwen2.5-VL-72B. 
Finally, when compared to leading closed-source reasoning models, the performance gap on a limited number of benchmarks such as Math-500, GSM8K, BBH remains close.

\FloatBarrier

\subsection{Evaluation Results in the Non-Thinking Mode}

For the evaluation under the non-thinking mode, we select a set of baseline models for comparison, including:
\begin{itemize}
    \item leading general-purpose models of similar scale, such as Qwen2.5-VL-3B\cite{bai2025qwen25vltechnicalreport}, Qwen3-4B-non-thinking~\cite{yang2025qwen3}, InternVL3-2B~\cite{zhu2025internvl3exploringadvancedtraining}, and Gemma-3-4B~\cite{gemmateam2025gemma3technicalreport}.
    \item larger-scale, high-impact general-purpose models, such as Kimi-VL-A3B-16B-non-thinking~\cite{kimiteam2025kimivltechnicalreport}.
    \item leading closed-source commercial models, such as OpenAI GPT-4.1\footnote{https://openai.com/index/gpt-4-1/}.
\end{itemize}


\subsubsection{Evaluation Results of Multimodal Benchmarks} 

As shown in Table~\ref{tab:multimodal-nothink}, compared to Qwen2.5-VL-3B, a model of similar scale (under 4B parameters), our model demonstrates superior performance across all metrics, with notable advantages on reasoning-related benchmarks such as MathVista and MathVision. 
Compared to Gemma-3-4B, BlueLM-2.5-3B leads in 9 out of 10 tasks.
When compared to larger-scale models such as Kimi-VL-A3B-16B-non-thinking, our model outperforms on more than half of the evaluation benchmarks, and for the remaining tasks except for MMMU, the performance gap remains within 5\%. 
Relative to closed-source commercial models, our model achieves results within 5\% on approximately half of the benchmarks such as MMBench-V1.1, OCRBench, AI2D, MathVista, HallusionBench.


\begin{table}[ht]
    \centering
    \caption{Comparison of BlueLM-2.5-3B-non-thinking and other baselines of Multimodal Benchmarks. The highest and second-best scores among open-source models are shown in \textbf{bold} and \underline{underline}, respectively.}
    \renewcommand{\arraystretch}{1.2} 
    \resizebox{0.98\textwidth}{!}{
    \begin{tabular}{l  C{3.0cm} C{2.5cm} C{2.5cm} | C{3.5cm} | C{2.5cm}}
        \toprule
        \textbf{} & \textbf{BlueLM-2.5-3B-non-thinking} & \textbf{Qwen2.5-VL-3B} & \textbf{Gemma-3-4B} & \textbf{Kimi-VL-A3B-16B-non-thinking} & \textbf{OpenAI-GPT-4.1} \\
        \midrule
        Params & 2.9B & 3.8B & 4.3B & A3B-16B & - \\
        \midrule
        MMMU & 47.5 & \underline{51.2} & 47.3 & \textbf{57.8} & 74.0 \\
        MMBench-V1.1 & \textbf{82.1} & 77.1 & 66.4 & \underline{80.8} & 86.6 \\
        MMstar & \textbf{64.5} & 56.5 & 47.9 & \underline{62.0} & 69.8 \\
        MM-vet & \textbf{66.7} & 63.2 & 57.8 & \underline{66.1} & 78.8 \\
        OCRBench & 82.6 & \underline{83.1} & 66.0 & \textbf{87.1} & 83.4 \\
        AI2D & \underline{83.0} & 81.4 & 70.7 & \textbf{84.5} & 85.9 \\
        MIA-bench & \underline{81.1} & 78.8 & \textbf{82.2} & 75.5 & 90.0 \\
        MathVista & \textbf{70.8} & 60.1 & 46.3 & \underline{66.0} & 70.4 \\
        MathVision & \textbf{28.5} & 21.2 & \underline{23.6} & 21.8 & 45.1 \\
        HallusionBench & \textbf{53.7} & 46.6 & 40.8 & \underline{48.4} & 58.5 \\
        \bottomrule
    \end{tabular}
    }
    \label{tab:multimodal-nothink}
\end{table}

\FloatBarrier

\subsubsection{Evaluation Results of Pure-Text Benchmarks} 

As shown in Table~\ref{tab:text-benchmark-nothink}, compared to Qwen2.5-VL-3B, our model demonstrates comprehensive performance advantages, with particularly notable improvements on reasoning-related benchmarks such as Math-500, BBH, AIME24, and AIME25. 
Compared to Gemma-3-4B, our model demonstrates superior performance across all benchmark datasets except IF-Eval, with substantial performance advantages observed in the majority of evaluation sets.
When compared to Qwen3-4B-non-thinking, the overall performance is comparable.
Relative to larger-scale models such as Kimi-VL-A3B-16B-non-thinking, our model outperforms across nearly all metrics, with especially strong advantages on reasoning-focused tasks.
However, there remains a measurable performance gap when compared with closed-source commercial models.


\begin{table}[ht]
    \centering
    \caption{Comparison of BlueLM-2.5-3B-non-thinking and other baselines of Text Benchmarks. The highest and second-best scores among open-source models are shown in \textbf{bold} and \underline{underline}, respectively.}
    \renewcommand{\arraystretch}{1.2} 
    \resizebox{0.99\textwidth}{!}{
    \begin{tabular}{l C{2.8cm} C{2.2cm} C{2.3cm} C{2.2cm} | C{3.5cm} | C{2.2cm}}
        \toprule
        \textbf{} & \textbf{BlueLM-2.5-3B-non-thinking} & \textbf{Qwen2.5-VL-3B} & \textbf{Qwen3-4B-non-thinking} & \textbf{Gemma-3-4B} & \textbf{Kimi-VL-A3B-16B-non-thinking} & \textbf{OpenAI-GPT-4.1} \\
        \midrule
        Params & 2.9B & 3.8B & 4.0B & 4.3B & A3B-16B & - \\
        \midrule
        MMLU-pro & \underline{60.2} & 37.4 & \textbf{61.3} & 43.6 & 44.2 & 81.0 \\
        GPQA-Diamond & \textbf{46.0} & 31.8 & \underline{41.7} & 30.8 & 33.3 & 69.2 \\
        Math-500 & \underline{80.0} & 58.0 & \textbf{84.8} & 75.6 & 56.6 & 91.8 \\
        GSM8K & \underline{90.1} & 77.7 & \textbf{90.8} & 89.2 &  78.8 & 95.7 \\
        BBH & \textbf{80.6} & 45.7 & \underline{79.5} & 72.2 &  54.3 & 87.5 \\
        AIME24 & \underline{13.3} & 3.3 & \textbf{25.0}& 6.7 &  6.7 & 50.0 \\
        AIME25 & \textbf{26.7} & 10.0 & \underline{19.1} & 6.7 & 0.0 & 30.0 \\
        IF-Eval & 78.4 & 46.6 & \underline{81.2} & \textbf{90.2} & 43.1 & 88.2 \\
        Humaneval & \textbf{85.4} & 71.3 & \underline{79.3} & 71.3 & 78.7 & 95.7 \\
        LiveCodeBench-v1 & \underline{43.0} & 15.5 & \textbf{48.0} & 9.5 & 22.8 & 68.2 \\
        LiveCodeBench-v5 & \textbf{21.7} & 5.4 & \underline{21.3} & 5.4 & 11.5 & 38.6 \\
        \bottomrule
    \end{tabular}
    }
    \label{tab:text-benchmark-nothink}
\end{table}

\FloatBarrier

\subsubsection{Evaluation Results of GUI Benchmarks} 

As shown in Table~\ref{tab:multimodal-ui-tars}, compared to the same scale model Qwen2.5-VL-3B, our model demonstrates comprehensive superiority in GUI grounding metrics, particularly on our in-house Chinese GUI grounding dataset ScreenSpot vivo. Meanwhile, BlueLM-2.5-3B's performance also surpasses that of UI-TARS-2B~\cite{DBLP:journals/corr/abs-2501-12326}, a domain model specialized for GUI applications, though a noticeable gap remains when compared to UI-TARS-7B~\cite{DBLP:journals/corr/abs-2501-12326}.


\begin{table}[ht]
    \centering
    \caption{Comparison of BlueLM-2.5-3B-non-thinking and other baselines of GUI Benchmarks. The highest and second-best scores are shown in \textbf{bold} and \underline{underline}, respectively.}
    \renewcommand{\arraystretch}{1.2} 
    \resizebox{0.98\textwidth}{!}{
    \begin{tabular}{l C{3cm} C{3cm} C{3cm} C{3cm} C{3cm}}
        \toprule
        \textbf{} & \textbf{BlueLM-2.5-3B} & \textbf{Qwen2.5-VL-3B} & \textbf{Qwen2.5-VL-7B} & \textbf{UI-TARS-2B} & \textbf{UI-TARS-7B} \\
        \midrule
        Params & 2.9B & 3.8B & 8.3B & 2.2B & 8.3B \\
        \midrule
        ScreenSpot & 82.7 & 55.5 & \underline{87.1} & 82.3 & \textbf{89.5} \\
        ScreenSpot V2 & \underline{85.8} & - & - & 84.7 & \textbf{91.6} \\
        ScreenSpot Pro & 28.7 & 23.9 & \underline{29.0} & 27.7 & \textbf{35.7} \\
        ScreenSpot vivo & \textbf{89.8} & - & 81.0 & - & \underline{85.8} \\
        \bottomrule
    \end{tabular}
    }
    \label{tab:multimodal-ui-tars}
\end{table}

\FloatBarrier

\section{Conclusion and Future Work} 

In this report, we present our efforts in building the multimodal BlueLM-2.5-3B model, which demonstrates strong
capabilities in both multimodal and pure-text reasoning, visual perception, and GUI agent grounding. 
Leveraging curated diverse pre-training datasets, key data resampling strategy, hybrid heterogeneous reinforcement learning, and high-performance training infrastructure, our model
demonstrates strong capability across evaluated benchmarks. We hope that our work advances the development of high-performance edge-side vision-language models and provides valuable insights for the community.
Our future work will expand the current modalities to train a single small-scale model capable of processing text, vision, and speech simultaneously. Moreover, we will incorporate more diversified reasoning data to continue enhancing the generalizability of the multimodal reasoning abilities.  

\FloatBarrier

\section*{Acknowledgements}

We thank the developers and maintainers of the open-source models and training frameworks that significantly supported our work. Specifically, we acknowledge Qwen, DeepSeek, and other open-source LLMs for providing strong baselines and synthetic data generation capabilities. We also appreciate the contributions of open training platforms such as Megatron-LM, DeepSpeed, and veRL, which enabled efficient large-scale model training.

\clearpage
\printbibliography[title={References}]

\newpage
\appendix
\section{Appendix}

\subsection{Contributors}

We would like to express our sincere gratitude to all contributors for their invaluable support and efforts. Authors within each role are listed in the alphabetical order of their first names.

\textbf{Contributors}

Baojiao Xiong, Boheng Chen, Chengzhi Wang, Daxiong Luo, Dongsheng Xu, Dongyang Liu, Fan Yang , Fangyuan Li, Fei Teng, Feng Wang , Fukang Qin, Fuquan Peng, Guanxin Tan, Guozhi Wang, Haibo Yu, Haohao Gao, Heng Liu, Hongbo Yang, Hongjian Zou, Houzheng Shen, Hu Meng, Huan Li, Hui Tan, Jiali Chen, Jianzhao Chen, Jinliang Zhu, Kai Wang, Lei Wu, Liangbing Liu, Liuyang Bian, Liyan He, Long Liu, Peiwen Li, Penggang Shi, Qi Ding, Rui Hu, Shuai Cao, Shuai Ren, Shuang Peng, Teng Xie, Weiji Chen, Weilin Xiang, Weixin Wu, Xi Yin, Xiaoxin Chen, Xu Chen, Yafei Wen, Yan Hu, Yanzhou Yang, Yina Xie, Yinghao Chen, Yixuan Liao, Yu Geng, Yuanjiang Ouyang, Yuanzhuo Yang, Yuehua He, Yushuai Peng, Zhaoxiong Wang, Zheng Wang, Zhibo Zhou, Ziyang Wu

\textbf{Acknowledgments}

We would like to sincerely thank Ao Tang, Bing Gong, Bo Xiao, Chang Hou, Fei Wu, Fei Xie, Hexiang Huang, Jidong Ye, Jie Wang, Liangbiao Pan, Min Chen, Min Jin, Pan Zhang, Panfeng Guo, Pengfei Su, Qiao Li, Renshou Wu, Ruijie Chen, Xiangyang Yu, Xinjie Huang, Yong Li, Yong Zhang, Yuhao Chen, Yuxiang Chai, Zeji Zhu, Zhibin Yao, Zhuang Jia, Zuguang Cai for their support to the project.

\subsection{Evaluation Benchmark List}
\label{app:bench_list}

\newpage
\begin{table}[h!]
    \centering
    \renewcommand{\arraystretch}{1.2}
    \caption{Multimodal Benchmarks}
    \begin{tabular}{l  m{11cm}}
    \toprule
        \textbf{Name} & \textbf{Introduction} \\
        \midrule
        \midrule
        MMMU~\cite{yue2024mmmu}  &  MMMU evaluates multimodal models on massive multi-discipline tasks, featuring 11.5K meticulously collected multimodal questions from college exams, quizzes, and textbooks, covering six core disciplines: Art \& Design, Business, Science, Health \& Medicine, Humanities \& Social Science, and Tech \& Engineering. \\
        \midrule
        MMBench~\cite{MMBench} &  MMBench is a vision-language benchmark developed by OpenCompass for evaluating multimodal models. MMBench contains approximately 3000 multiple-choice questions spanning 20 ability dimensions, systematically testing abilities from basic perception to advanced cognition. \\
        \midrule
        MMStar~\cite{chen2024mmstar} & MMStar is a multi-modal benchmark comprising 1,500 samples meticulously selected by humans. MMStar benchmarks 6 core capabilities and 18 detailed axes, aiming to evaluate LVLMs' multi-modal capacities with carefully balanced and purified samples. \\
        \midrule
        MM-vet~\cite{yu2024mmvet} & MM-Vet is an evaluation benchmark that examines large multimodal models (LMMs) on complicated multimodal tasks. It systematically assesses 16 functional combinations spanning 6 core vision-language (VL)  capabilities. \\
        \midrule
        OCRBench~\cite{ocrbench} & OCRBench contains 29 datasets, including Text Recognition, Scene Text-Centric Visual Question Answering (VQA), Document-Oriented VQA, Key Information Extraction (KIE), and Handwritten Mathematical Expression Recognition (HMER), making it the most comprehensive OCR evaluation benchmark. \\
        \midrule
        AI2D~\cite{kembhavi2016ai2d} & AI2 Diagrams (AI2D) is a dataset of over 5000 grade school science diagrams with over 150000 rich annotations, their ground truth syntactic parses, and more than 15000 corresponding multiple-choice questions. \\
        \midrule
        MIA-bench~\cite{qian2025miabenchbetterinstructionfollowing} & MIA-bench evaluates multimodal large language models (MLLMs) on their ability to adhere to complex instructions, comprising a diverse set of 400 image-prompt pairs. \\
        \midrule
        MathVista~\cite{lu2023mathvista} & MathVista is designed to assess the mathematical capabilities of multimodal large models by combining challenges from diverse mathematical and visual tasks. The benchmark consists of 6,141 examples, derived from 28 existing multimodal datasets involving mathematics and 3 newly created datasets. \\
        \midrule
        MathVision~\cite{wang2024measuring} & MathVision is designed to evaluate the mathematical reasoning capabilities of multimodal large models. It comprises 3,040 high-quality math problems spanning 16 distinct mathematical disciplines and graded across 5 levels of difficulty. \\
        \midrule
        HallusionBench~\cite{guan2024hallusionbenchadvanceddiagnosticsuite} & HallusionBench is a comprehensive benchmark designed for the evaluation of image-context reasoning, comprising 346 images paired with 1129 questions. \\
       \bottomrule
    \end{tabular}
    \label{tab:multimodal_intro}
\end{table}

\newpage
\begin{table}[h!]
    \centering
    \renewcommand{\arraystretch}{1.3}
    \begin{threeparttable}
    \caption{Pure-Text Benchmarks}
    \label{tab:pure_text_intro}
    \begin{tabular}{l  m{11cm}}
    \toprule
        \textbf{Name} & \textbf{Introduction} \\
        \midrule
        MMLU-pro~\cite{wang2024mmluprorobustchallengingmultitask} & MMLU-Pro, an extended version of MMLU, introduces more challenging reasoning-focused questions while expanding the choice set from four to ten options. \\
        \midrule
        GPQA-Diamond~\cite{rein2023gpqagraduatelevelgoogleproofqa} & GPQA Diamond, the highest-quality subset of the GPQA series, consists of 198 questions, while the standard GPQA includes 448 questions. This benchmark specifically evaluates a model's ability to solve problems requiring expert-level understanding and reasoning. \\
        \midrule
        Math-500~\cite{lightman2023lets} & The MATH-500 benchmark was launched by OpenAI in 2023 as a tool for evaluating the mathematical abilities of its latest models, such as GPT-4o. This benchmark features 500 high-difficulty mathematical competition problems, specifically designed to challenge the limits of these models and assess their reasoning and problem-solving capabilities in complex mathematical scenarios. \\
        \midrule
        GSM8K~\cite{cobbe2021gsm8k} & GSM8K is a dataset containing 8.5K high-quality, linguistically diverse grade school math problems created by human problem writers. The dataset is segmented into 7.5K training problems and 1K test problems. These problems take between 2 and 8 steps to solve, and solutions primarily involve performing a sequence of elementary calculations to reach the final answer. \\
        \midrule
        BBH~\cite{suzgun2022challengingbigbenchtaskschainofthought} & BIG Bench-Hard (BBH) is a subset of BIG Bench, a diverse evaluation suite for language models. BBH focuses on 23 challenging tasks from BIG Bench that have been found to be beyond the capabilities of current language models. \\
        \midrule
        AIME24 & The dataset contains 30 mathematical problem sets from the 2024 AIME Dataset. \\
        \midrule
        AIME25 & The dataset contains 30 mathematical problem sets from the 2025 AIME Dataset. \\
        \midrule
        IF-Eval~\cite{Zhou2023InstructionFollowingEF} & IFEval is a straightforward and easy-to-reproduce evaluation benchmark. It focuses on a set of "verifiable instructions" such as "write in more than 400 words" and "mention the keyword of AI at least 3 times". \\
        \midrule
        Humaneval*~\cite{chen2021evaluatinglargelanguagemodels} & HumanEval consists of 164 original  programming problems, assessing language comprehension, algorithms, and simple mathematics, with some comparable to simple software interview questions. \\
        \midrule
        LiveCodeBench~\cite{Jain2024LiveCodeBenchHA} & LiveCodeBench is a benchmark for evaluating the code-related capabilities of large language models (LLMs), continuously collecting new problems from LeetCode, AtCoder, and CodeForces. Particularly, LiveCodeBench also focuses on broader code-related capabilities. \\
        \bottomrule
    \end{tabular}
    \begin{tablenotes}
        \item * Humaneval uses Pass@1 evaluation metric.
    \end{tablenotes}
    \end{threeparttable}
    
\end{table}





\newpage
\subsection{Qualitative Examples}

\begin{figure}[h]
\centering
\renewcommand{\arraystretch}{1.75}
\begin{tabular}{m{15cm}}
\toprule
\textbf{Input Image} \\

\begin{center}
\includegraphics[width=0.6\textwidth]{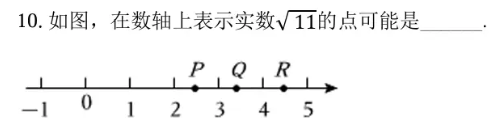}
\end{center} \\

\textbf{Prompt: } \begin{CJK*}{UTF8}{gbsn}回答图中的问题\end{CJK*} \\

\textbf{Response: } 
\begin{CJK*}{UTF8}{gbsn}

\texttt{<think>}
首先，我需要确定$\sqrt{11}$的大致数值范围。已知$3^2$等于9，$4^2$等于16，而11位于9和16之间。因此，$\sqrt{11}$应该在3到4之间。接下来，观察数轴上的点P、Q和R的位置。点P大约在2.5的位置，点Q在3附近，点R则在4左右。根据计算结果，$\sqrt{11}$大约为3.3166，所以它应该靠近点Q。
\texttt{</think>}
\newline
要确定$\sqrt{11}$在数轴上对应的点$P$、$Q$或$R$，我们可以按照以下步骤进行分析:

1. **估算$\sqrt{11}$的范围**:
   - 已知$3^2 = 9$ 和 $4^2 = 16$。
   - 因此，$\sqrt{11}$位于3和4之间，即$3 < \sqrt{11} < 4$。

2. **比较各点的位置**：
   - 点$P$位于2和3之间，显然小于$\sqrt{11}$。
   - 点$Q$位于3和4之间，且更接近3。由于$\sqrt{11} \approx 3.3166$，点$Q$正好落在这一区间内。
   - 点$R$位于4附近，明显大于$\sqrt{11}$。

综上所述，表示实数$\sqrt{11}$的点可能是\boxed{Q}
\end{CJK*} \\

\bottomrule
\end{tabular}

\caption{Example of BlueLM-2.5-3B solving mathematical OCR problems with reasoning.}
\label{fig:math ocr}
\end{figure}

\begin{figure}
\centering
\renewcommand{\arraystretch}{1.75}
\begin{tabular}{m{15cm}}
\toprule
\textbf{Input Image} \\

\begin{center}
\includegraphics[width=0.8\textwidth]{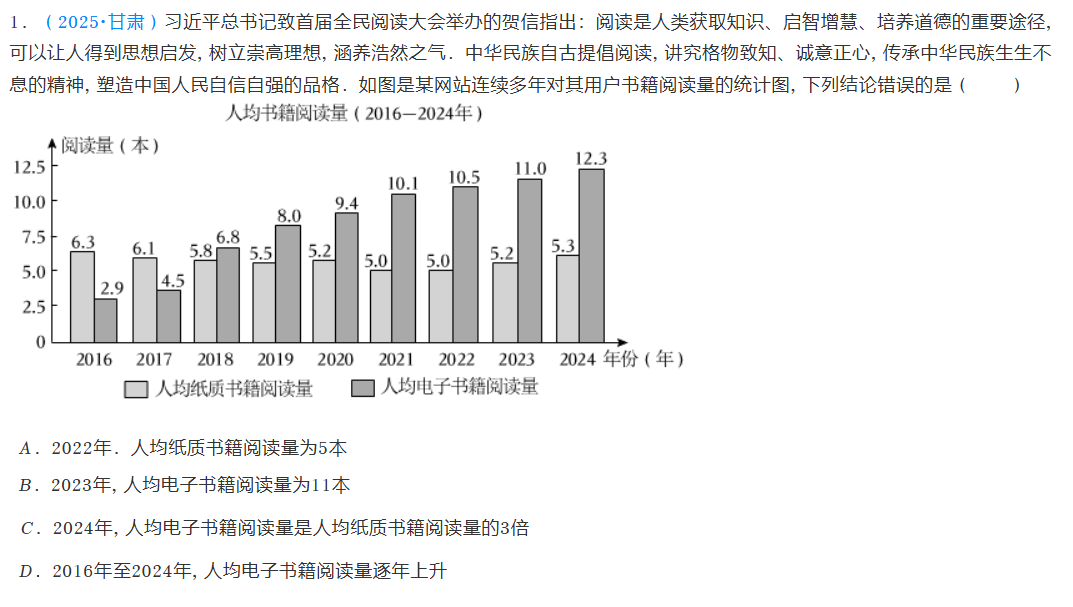}
\end{center} \\

\textbf{Prompt: } \begin{CJK*}{UTF8}{gbsn}回答图中选择题\end{CJK*} \\

\textbf{Response: } 
\begin{CJK*}{UTF8}{gbsn}

\texttt{<think>}$\dots$\texttt{</think>}
\newline
**分析与解答过程：**

根据图表数据，逐一验证各选项的正确性：

- **选项A**：2022年，人均纸质书籍阅读量为5本。  
  图表中显示2022年的纸质书阅读量为5.0本，与选项一致，故A正确。

- **选项B**：2023年，人均电子书籍阅读量为11本。  
  图表中显示2023年的电子书阅读量为11.0本，与选项一致，故B正确。

- **选项C**：2024年，人均电子书籍阅读量是人均纸质书籍阅读量的3倍。  
  计算2024年的比例：\( \frac{12.3}{5.3} \approx 2.32 \)，并非3倍，故C错误。

- **选项D**：2016年至2024年，人均电子书籍阅读量逐年上升。  
  图表中每年电子书阅读量均高于前一年，故D正确。

**最终答案：**  \boxed{C}
\end{CJK*} \\

\bottomrule
\end{tabular}

\caption{Example of BlueLM-2.5-3B solving bar-chart OCR comprehension problems.}
\label{fig:bar-chart OCR}
\end{figure}

\begin{figure}
\centering
\renewcommand{\arraystretch}{1.75}
\begin{tabular}{c m{8cm}}
\toprule
 \multirow{2}{*}{
 \includegraphics[width=0.4\textwidth]{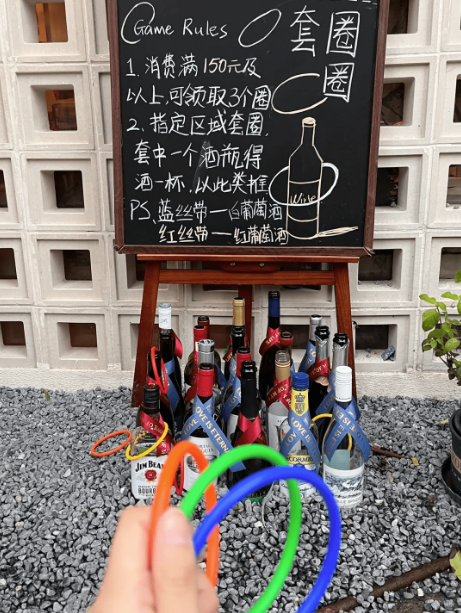}
 } & \textbf{Prompt: } How many circles are there in the person’s hand in the picture? \\
 
 & \textbf{Response: } 
 \begin{CJK*}{UTF8}{gbsn}
    \texttt{<think>}
用户现在需要数图片中手里的圆环数量。看图片里，手里拿着的圆环有橙色、绿色、蓝色？不对，看清楚：应该是三个？不，数一下。手里拿着的是橙色、绿色、蓝色？不对，仔细看，是三个吗？哦，不对，看颜色：橙色、绿色、蓝色？不，重新数。图中手拿着的圆环，分别是橙色、绿色、蓝色？不对，应该是三个？不对，再看。哦，对，有三根圆环？不，不对，看图片里的圆环：橙色、绿色、蓝色，这是三个？不对，再数。哦，对，手里拿着的三个圆环？不对，不对，看清楚。哦，用户手中的圆环，橙色、绿色、蓝色，所以是3个？不对，再看。哦，对，图中有三个圆环（三根不同颜色的圈）。所以答案是3。
\texttt{</think>}

 \end{CJK*}

 To determine the number of circles (hoop rings) in the person's hand, we count the hoop rings visible in the image. There are three hoop rings (one orange, one green, and one blue).

\boxed{3} \newline \\




\midrule
\midrule

 \multirow{2}{*}{
 \includegraphics[width=0.4\textwidth]{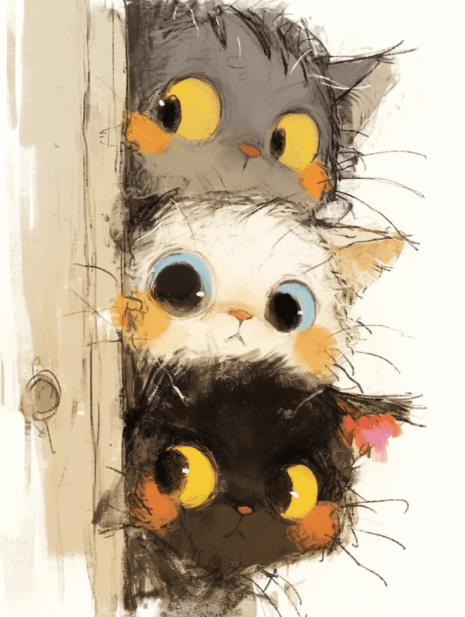}
 } & \textbf{Prompt: } What colors are the three cats in the picture from top to bottom? \\
 
 & \textbf{Response: } 

\texttt{<think>}

So, let's look at the image. The first cat at the top is gray. Then the middle one is white with orange patches. The bottom cat is black. So from top to bottom, the colors are gray, white, black.

\texttt{</think>}
\newline
To determine the colors of the three cats from top to bottom, we analyze the image:  
1. The top - most cat has a gray color.  
2. The middle cat is white (with orange patches).  
3. The bottom - most cat is black.  

Thus, the colors from top to bottom are gray, white, and black.
\newline
\newline
\newline
\newline
\newline
\newline

\\

\bottomrule
\end{tabular}

\caption{Example of BlueLM-2.5-3B solving visual perception problems with reasoning.}
\label{fig:visual reasoning}
\end{figure}

This section presents a selection of qualitative examples illustrating the capabilities of the BlueLM-2.5-3B model
through responses generated from various prompts. Examples demonstrating the model’s ocr reasoning capacity are provided in Figure~\ref{fig:math ocr} and Figure~\ref{fig:bar-chart OCR}. Figure~\ref{fig:visual reasoning} illustrates the model's visual perception capability. Figure~\ref{fig:text reasoning} demonstrates the model's text reasoning capability. Finally, Figure~\ref{fig:gui grounding} showcases the model's GUI agentic ability.

\newpage

\begin{figure}
\centering
\renewcommand{\arraystretch}{1.75}
\begin{tabular}{m{15cm}}
\toprule

\textbf{Prompt: } \begin{CJK*}{UTF8}{gbsn}已知$53\triangle37=315$，$88\triangle32=384$，那么$27\triangle36=?$\end{CJK*} \\

\textbf{Response: } 
\begin{CJK*}{UTF8}{gbsn}

\texttt{<think>}$\dots$\texttt{</think>}
\newline
已知$53\triangle37=315$，$88\triangle32=384$，要求求出$27\triangle36$的值。我们分析得出以下结论：

**运算符△的定义**  
该运算符△的作用是将第一个数的十位与第二个数的个位相乘，并将第一个数的个位与第二个数的十位相乘，最后将这两个乘积相乘。具体步骤如下：
1. 将第一个数的十位与第二个数的个位相乘；
2. 将第一个数的个位与第二个数的十位相乘；
3. 将上述两步的结果相乘，即为最终结果。

**应用定义求解27△36**  
对于27△36：
- 第一个数27的十位是2，个位是7；
- 第二个数36的十位是3，个位是6。

按步骤计算：
1. \(2 \times 6 = 12\)（第一个数的十位与第二个数的个位相乘）；
2. \(7 \times 3 = 21\)（第一个数的个位与第二个数的十位相乘）；
3. 最终结果：\(12 \times 21 = 252\)。

最终答案为 \(\boxed{252}\)。

\end{CJK*} \\




\midrule
\midrule

\textbf{Prompt: } \begin{CJK*}{UTF8}{gbsn}爱丽丝有N个兄弟，她还有M个姐妹。爱丽丝的兄弟有多少个姐妹？\end{CJK*} \\

\textbf{Response: } 
\begin{CJK*}{UTF8}{gbsn}

\texttt{<think>}$\dots$\texttt{</think>}
\newline
要解决这个问题，我们首先要明确几个基本概念：

一、变量定义

- **N** 表示爱丽丝的兄弟人数；
- **M** 表示爱丽丝的姐妹人数（不包括爱丽丝本人）；
- 问题要求求出：**爱丽丝的兄弟有多少人作为姐妹**。

二、核心逻辑

我们要找的是：**爱丽丝的兄弟所看到的“姐妹”的人数**。换句话说，这是从兄弟视角出发，统计他们家有多少个女性成员（即姐妹）。

 分析步骤如下：

1. **姐妹总数**：
   - 如果爱丽丝有 $ M $ 个姐妹（不包括她自己），那么整个家庭中就有 $ M + 1 $ 个姐妹（包括爱丽丝）。

2. **兄弟视角下的姐妹数**：
   - 每个兄弟都会看到全部的家庭女性成员，因为这些女性都不是他/她的兄弟。
   - 所以，无论哪个兄弟看待，他都会看到 $ M + 1 $ 个姐妹。

三、示例验证

假设：
- 爱丽丝有 2 个兄弟（$ N = 2 $）；
- 她有 3 个姐妹（$ M = 3 $，不包括她自己）。

那么：
- 整个家庭中共有 $ M + 1 = 4 $ 个姐妹（包括爱丽丝）；
- 每个兄弟也会看到这 4 个姐妹。

四、最终答案

$$
\text{爱丽丝的兄弟有 } \boxed{M + 1} \text{ 个姐妹。}
$$

\end{CJK*} \\

\bottomrule
\end{tabular}

\caption{Example of BlueLM-2.5-3B solving complex text reasoning problems.}
\label{fig:text reasoning}
\end{figure}

\begin{figure}
\centering
\renewcommand{\arraystretch}{1.75}
\begin{tabular}{m{15cm}}
\toprule
\textbf{Task: } \begin{CJK*}{UTF8}{gbsn}帮我搜索头戴式的蓝牙耳机，并筛选出价格区间在0到300的。\end{CJK*} \\

\begin{center}
\includegraphics[width=0.9\textwidth]{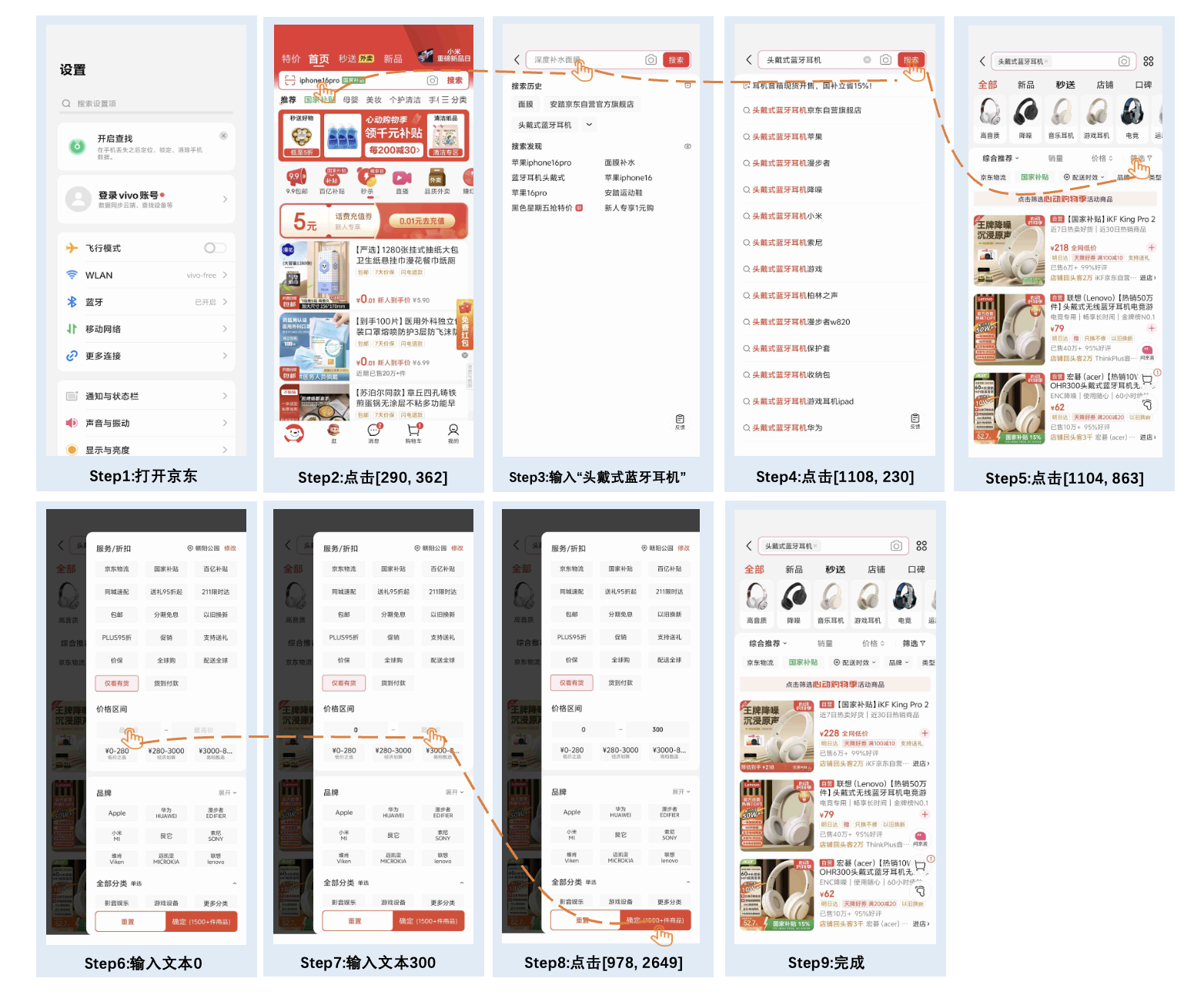}
\end{center} \\

\bottomrule
\end{tabular}

\caption{Example of BlueLM-2.5-3B solving GUI agent problems.}
\label{fig:gui grounding}
\end{figure}

\end{document}